\pdfoutput=1 % ensure pdflatex (for arXiv)

\documentclass[11pt]{article}
\usepackage[]{acl2023}
\usepackage{nert}
\usepackage{times}
\usepackage{latexsym}
\usepackage{amssymb}% http://ctan.org/pkg/amssymb
\usepackage{pifont}% http://ctan.org/pkg/pifont
\newcommand{\cmark}{\ding{51}}%
\newcommand{\xmark}{\ding{55}}%
\usepackage[inline,shortlabels]{enumitem}
\usepackage{comment}
\usepackage{multirow}
\usepackage{hyperref}

\usepackage{todonotes}
\usepackage{microtype}
\usepackage{xcolor}

\usepackage{kotex}

\newcounter{mycounter}

\title{BLOOM+1: Adding Language Support to BLOOM for Zero-Shot Prompting}
        \author{
\textbf{Zheng-Xin Yong}$^{1}$\thanks{\ \ Corresponding author: contact.yong@brown.edu}\ \ , 
\textbf{Hailey Schoelkopf}$^{2, 3}$, 
\textbf{Niklas Muennighoff}$^{4}$,  
\textbf{Alham Fikri Aji}$^{5}$,
\\
\textbf{David Ifeoluwa Adelani}$^{6}$, 
\textbf{Khalid Almubarak}$^{7}$, 
\textbf{M Saiful Bari}$^{8}$, 
\textbf{Lintang Sutawika}$^{2, 9}$, 
\\
\textbf{Jungo Kasai}$^{10}$,
\textbf{Ahmed Baruwa}$^{11}$,
\textbf{Genta Indra Winata}$^{12}$,
\textbf{Stella Biderman}$^{2, 13}$,
\\
\textbf{Edward Raff}$^{13}$,
\textbf{Dragomir Radev}$^{3}$,
\textbf{Vassilina Nikoulina}$^{14}$
\\
$^1$Brown University \;
$^2$EleutherAI \;
$^3$Yale University \;
$^4$Hugging Face \; 
$^5$MBZUAI \; 
\\ 
$^6$University College London \; 
$^7$PSAU \; 
$^8$Nanyang Technological University \;
\\ 
$^9$Datasaur.ai \; 
$^{10}$Paul G.\ Allen School of CSE, University of Washington \; 
\\
$^{11}$University of Oregon \;
$^{12}$The Hong Kong University of Science and Technology \; 
\\
$^{13}$Booz Allen Hamilton \; 
$^{14}$NAVER LABS Europe
}

\begin{document}
\maketitle
\begin{abstract}
The BLOOM model is a large publicly available multilingual language model, but its pretraining was limited to 46 languages. To extend the benefits of BLOOM to other languages without incurring prohibitively large costs, it is desirable to adapt BLOOM to new languages not seen during pretraining. In this work, we apply existing language adaptation strategies to BLOOM and benchmark its zero-shot prompting performance on eight new languages in a \textit{resource-constrained} setting. We find language adaptation to be effective at improving zero-shot performance in new languages. Surprisingly, we find that adapter-based finetuning is more effective than continued pretraining for large models. In addition, we discover that prompting performance is not significantly affected by language specifics, such as the writing system. It is primarily determined by the size of the language adaptation data. We also add new languages to BLOOMZ, which is a multitask finetuned version of BLOOM capable of following task instructions zero-shot. We find including a new language in the multitask fine-tuning mixture to be the most effective method to teach BLOOMZ a new language. We conclude that with sufficient training data language adaptation can generalize well to diverse languages. Our code is available at \url{https://github.com/bigscience-workshop/multilingual-modeling}.

\end{abstract}
\section{Introduction}
Although access to transformer-based language models has expanded greatly over the past several years \citep{black2021gpt,wang2021gpt,artetxe2021efficient,black-etal-2022-gpt,zhang2022opt}, these technologies are overwhelmingly concentrated in a few high resource languages \citep{talat-etal-2022-reap}. BLOOM \cite{scao-2022-bloom}, the largest publicly available multilingual language model to date with 176B parameters, covers only 46 natural languages and even excludes high-resource languages such as Korean and Russian which has tens of millions of speakers.
%Current multilingual language models have limited coverage of languages in their pretraining data. For instance, BLOOM \cite{scao-2022-bloom}, which is the largest multilingual open-source model to date created by the BigScience community, only covers 46 natural languages.
This limitation was driven by a number of factors, most notably only considering languages for which the community had enough expertise to manually validate the data quality \citep{kreutzer-2022-quality}, deduplicate and remove personally identifiable information \cite{laurencon-2022-roots} and had sufficient access to licensed unlabeled text \cite{joshi-etal-2020-state}. All of these factors are contingent facts about the group that trained the model, and leave open the idea that other researchers could contribute more languages. As regularly retraining such a model is prohibitively expensive, the question of whether this model can be productively \textit{adapted} to understand additional languages after training becomes pressing.

We hypothesize that language adaptation scenario is especially interesting for low-resource languages that would benefit from knowledge transfer. Therefore, we adapt BLOOM models to support eight new languages (German, Russian, Bulgarian, Thai, Turkish, Greek, Korean, and Guarani) in the resource-constrained settings, where we only use a limited amount of samples (maximum 100K samples) for each language. We  evaluate their zero-shot prompting on various NLU tasks after adaptation. The new languages cover both seen and unseen scripts in the pretraining data, and they differ in their language families and word orders. We benchmark existing language adaptation methods, such as continued pretraining and MAD-X \cite{pfeiffer-etal-2020-mad}, as well as a state-of-the-art parameter-efficient transfer learning method, (IA)$^3$ \cite{liu2020tfew}.

Current work on adapting large multilingual models has mostly explored continued pretraining \cite{muller2022cedille,genji-gpt-j,de2022zero} of EleutherAI's GPT-J-6B \citep{wang2021gpt}. Moreover, \citet{ebrahimi-kann-2021-adapt} showed that continued pretraining outperforms other strategies for adapting small/medium-sized language models (i.e., models with fewer than one billion parameters).
However, our experiments demonstrate that, for large language models such as BLOOM with comparable sizes to GPT-J-6B, continued pretraining underperforms adapters under a resource-constrained setting. In addition, our work focuses on studying the effects of language adaptation on prompting, which has been underexplored in previous language adaptation work \cite{ebrahimi-kann-2021-adapt,ansell-etal-2022-composable,parovic-etal-2022-bad,pfeiffer-etal-2022-lifting}. Prompting can benefit many languages that lack large amounts of labeled data as it allows language models to generalize to a wide range of tasks with significantly less training cost and data than full finetuning \cite{liu2021promptsurvey,le-scao-rush-2021-many}. 

% While previous work has investigated different language adaptation strategies \cite{pfeiffer-etal-2020-mad,ansell-etal-2022-composable,parovic-etal-2022-bad}, most literature studies small language models such as mBERT \cite{devlin-etal-2019-bert,pires-etal-2019-multilingual} and XLM-Roberta \cite{conneau-etal-2020-unsupervised}. Our work shows that some previous findings \cite{ebrahimi-kann-2021-adapt} do not necessarily generalize when language models scale up. 
% Furthermore, thus far limited work has explored the effects of language adaptation on prompting, which allows language models to generalize to a wide range of tasks with significantly less training cost and data than full finetuning \cite{liu2021promptsurvey,le-scao-rush-2021-many}. Studying the relationship between language adaptation and prompting can hugely benefit many languages that lack large amounts of labeled data.

\subsection{Our Contributions}

Our work is the first to explore the \textit{scaling effects} of \textit{language adaptation} strategies for language models with billions of parameters under a \textit{resource-constrained} setting. 
Contrary to prior work on small/medium-sized multilingual masked language models \cite{ebrahimi-kann-2021-adapt}, we recommend training adapters instead of continued pretraining for BLOOM with at least 3 billion parameters for better prompting performance. We further connect this recommendation to the way the quality of language independent representation scales with model parameters.

We also demonstrate the positive effects of monolingual language adaptation on the prompting performance of BLOOM on various datasets. BLOOMZ is a variant of BLOOM that is produced by finetuning BLOOM on a multitask mixture in the same languages seen during pretraining. We find that simply adding a new language in the multitask finetuning is effective in improving performance in the new language.

To summarize, our contributions include:
\begin{itemize}
    \item Studying the effects of language adaptation on zero-shot prompting and instruction tuning.
    \item Benchmarking parameter-efficient methods for adapting BLOOM models of various scales and analyzing the trade-offs between the amount of required computes and zero-shot prompting performance.
    \item Quantifying the effect of the size of language adaptation data on language adaptation.
\end{itemize}

\section{Related Work}
\paragraph{Language Adaptation} Language adaptation enables pretrained language models to support languages outside of their pretraining data. Most works investigating language adaptation consider masked language models such as mBERT \cite{devlin-etal-2019-bert} and XLM-R \cite{conneau-etal-2020-unsupervised} that are pretrained on 100+ languages. Language adaptation approaches can be broadly categorized into three categories: (1) \textit{continued pretraining} of the model (restricted to the embedding layer training only in some cases) \cite{neubig-hu-2018-rapid,artetxe-etal-2020-cross,chau-etal-2020-parsing,muller-etal-2021-unseen,zhang-etal-2020-multi-stage,wang-etal-2020-extending}; (2) training of \textit{language-specific adapters} \cite{pfeiffer-etal-2020-mad, pfeiffer-etal-2021-adapterfusion, pfeiffer-etal-2021-unks, philip-etal-2020-monolingual, ustun-etal-2021-multilingual, berard-2021-continual, faisal-anastasopoulos-2022-phylogeny,parovic-etal-2022-bad} for the target language; and (3) training of a \textit{sparse subset} of model parameters \cite{ansell-etal-2022-composable}.
The core motivation behind these approaches is to benefit from knowledge transfer encoded in the pretrained language models for the new language processing at a small computational cost (compared to full model retraining from scratch).

One common issue is that the script of the new language is not always supported by the tokenizer. \citet{artetxe-etal-2020-cross, aji2020neural, pfeiffer-etal-2021-unks} demonstrate that it is possible to add a new language to these models by training a new embedding layer. \citet{muller-etal-2021-unseen} continue training the pretrained mBERT on the new language data, and find that transliteration of languages using non-Latin script boosts performance on these languages.
\citet{berard-2021-continual} add new languages into pretrained multilingual machine translation models by training embedding and adapter layers. They show that adding a new target language (the language to translate to) is harder to learn than a new language to translate from.

%\cite{arr_jan2022} proposes to revise the architecture of pretrained multilingual models by adding an explicit language-specific module. They demonstrate that such architecture suffers less from \textit{curse of multilinguality} and allows to easily add new language to the pretrained model by training new language-specific module. 

Closest work to our benchmarking efforts is \citeposs{ebrahimi-kann-2021-adapt} study on different approaches (i.e., continued pretraining, vocabulary expansion and adapter layers) to extend the XLM-R model to 30 new languages on token-level classification tasks.
They conclude that continued pretraining is the most promising direction. However, the cost of such pretraining will grow with the size of the pretrained model and can be prohibitive for many researchers working with low-resource languages.
Our results also show that continued pretraining does not necessarily bring a prompting performance gain for larger language models.

% Our work aims to compare lightweight strategies  of adaptation to new language (embedding layer training and adapter layers) across the following previously unstudied dimensions: we consider the (1) adaptation parameter budget and (2) intermediate pretraining checkpoints by analyzing how they impact autoregressive model's adaptability to new languages.

\paragraph{Multilingual Prompting} Prompting reformulates NLP tasks into masked or generative language modeling problem, depending on the models' pretraining objective. \citet{zhao-schutze-2021-discrete} and \citet{qi-etal-2022-enhancing} show that finetuning XLM-R on cloze-style prompts yield better performance than standard finetuning under a low-resource regime for XNLI. On the other hand, \citet{winata-etal-2022-cross} find that standard finetuning of XLM-R outperforms prompt-based learning for sentiment prediction in low-resource Indonesian dialects.

Some work shows that multitask prompt-based training on a variety of tasks and English or translated prompts improves zero-shot cross-lingual and cross-task performance~\cite{muennighoff2022bloomz,fu2022polyglot}. Multilingual prompt-based learning can also be achieved without performing gradient updates for downstream tasks. For instance, \citet{xi-2021-xglm} demonstrate success in prompting GPT-like pretrained models with in-context learning for NLU tasks, using either English or translated prompt templates. \citet{shi2023language} find that when language models scale up, they can perform better multilingual chain-of-thought reasoning. 
\section{Experimental settings}
\subsection{BLOOM pretrained models} 
We focus on adding language support to the BLOOM language model \cite{scao-2022-bloom} from 560 million to 7.1 billion parameters. BLOOM has a decoder-only Transformer architecture that uses AliBi positional embeddings \citep{press-2022-alibi} and layer normalization after embedding layers. Its tokenizer is trained with byte-level Byte Pair Encoding (BPE) algorithm \cite{gage-1994-bpe, sennrich2016neural} with a vocabulary size of 250,680. 

BLOOM is pretrained for around 350 billion tokens on the ROOTS corpus \cite{laurencon-2022-roots}, which covers 46 natural languages and 13 programming languages. Appendix~\ref{app:seen-lang} shows the distribution of the natural languages in the ROOTS corpus. 

\subsection{New Languages}
We consider all six languages of XNLI~\cite{conneau-etal-2018-xnli} that are currently unsupported by BLOOM: German, Bulgarian, Russian, Greek, Turkish, and Thai. We also include Korean to follow up on past work on adapting the previous version of BLOOM \cite{yong2022adapting} and Guarani, which is a truly low-resource Native American language. Table~\ref{tab:languages} summarizes the unseen languages used in our experiments. They cover different language families and some of them do not share scripts with BLOOM's supported languages. 

\begin{table*}[ht]
\small
    \centering
    \begin{tabular}{llllcc}
        \toprule
        Language & Language Family & Word Order & Script & Space-Separated & Seen Script \\
        \midrule
        German & Indo-European (Germanic) & SVO & Latin & \cmark & \cmark \\
        Bulgarian & Indo-European (Slavic) & SVO & Cyrillic & \cmark & \xmark \\
        Russian & Indo-European (Slavic) & SVO & Cyrillic & \cmark & \xmark \\
        Greek & Indo-European (Hellenic) & SVO & Greek & \cmark & \xmark \\
        Turkish & Turkic & SOV & Latin & \cmark & \cmark \\
        Korean & Koreanic & SOV & Hangul & \cmark & \xmark \\
        Thai & Tai–Kadai & SVO & Thai & \xmark & \xmark \\
        Guarani & Tupian & SVO & Latin & \cmark & \cmark \\
        \bottomrule
    \end{tabular}
    \caption{Information about the unseen languages used in our experiments. 
    %We report the contamination of the unseen languages in the 1\% of randomly sampled documents from ROOTS \cite{laurencon-2022-roots,muennighoff2022bloomz}.
    }
    \label{tab:languages}
\end{table*}

%%%%%%%%% language adapation strategies %%%%%%%%%
\subsection{Language Adaptation Strategies} \label{sec:lang-adapt-strats}
We carry out three language adaptation strategies to analyze their effects on zero-shot prompting. \footnote{We also ran preliminary experiments on Composable Sparse-Finetuning (see Appendix~\ref{app:c-sft}), which is one of the state-of-the-art language adaptation strategies.}

\paragraph{Continued Pretraining} Continued pretraining strategy refers to continually training the BLOOM model with its causal language modeling pretraining objective on monolingual text of the new language \cite{chau-etal-2020-parsing,ebrahimi-kann-2021-adapt,muller-etal-2021-unseen}. 

\paragraph{MAD-X} We use the language adapter and the invertible adapter of the MAD-X configuration \cite{pfeiffer-etal-2020-mad} to adapt BLOOM to new languages. Language adapter refers to the bottleneck adapter with down- and up-projection feedforward layers \cite{houlsby2019parameter,pfeiffer-etal-2021-adapterfusion} that are inserted into each Transformer block.
The invertible adapter is used in the embedding layers to mitigate the mismatch between the original and new language vocabularies. 

\paragraph{(IA)$^3$} 
(IA)$^3$ is a parameter-efficient finetuning method that performs element-wise rescaling of inner Transformer block activations 
 through learnable vectors~\cite{liu2020tfew}. These vectors can be merged with the original pretrained weights of a model at inference to reduce latency by avoiding passing the activations through additional adapter modules.

We experiment with (IA)$^3$ since it outperforms bottleneck adapters, which are used in MAD-X, and other parameter-efficient finetuning methods such as BitFit \cite{ben-zaken-etal-2022-bitfit}, LoRA \cite{hu2022lora}, and FishMask \cite{sung2021training} on English NLU tasks \cite{liu2020tfew}. Our preliminary experiments show that (IA)$^3$ performs better than these methods (see Appendix~\ref{app:lang-adapt-strats}), and thus we only run (IA)$^3$ due to computational constraints.

As (IA)$^3$ does not adapt the embedding layer, we couple (IA)$^3$ with invertible adapters for fairer comparison with MAD-X language adapters. Our preliminary experiments (Table~\ref{tab:lang-adapt-strats}) show performance gains when using invertible adapters with (IA)$^3$.

%%%%%%%%% adaptation setting %%%%%%%
\subsection{Language Adaptation Setting}
\label{sec:lang-adapt-setting}
We randomly sample 100K samples from the deduplicated OSCAR subcorpora \citep{ortiz:oscar_2019} of the respective languages for language adaptation to simulate low-resource settings. Since Guarani only has around 100 samples in OSCAR, we use Jojajovai parallel corpora \cite{chiruzzo-etal-2022-jojajovai}, which contains 30K Guarani sentences. We perform 25K language adaptation training steps using a batch size of 8 and the sequence length of 1,024. See Appendix~\ref{app:lang-adapt-details} for further details.

We do not retrain the tokenizer as BLOOM uses byte-level BPE tokenization, which never produces unknown tokens; therefore, we can perform language adaptation \textit{without} extending the vocabulary. We adapt the embedding layer in two different fashions. For continued pretraining, we make the embedding layer trainable. This follows prior work on language adaptation \cite{pfeiffer-etal-2020-mad,chau-etal-2020-parsing,ebrahimi-kann-2021-adapt,fujinuma-etal-2022-match}. For MAD-X and (IA)$^3$, we use invertible adapters to adapt the embedding layer while keeping the embeddings frozen.

%%%%% Evaluation Task
\subsection{Tasks and Prompt Templates}
We evaluate the models on five multilingual NLU tasks, which cover natural language inference (XNLI \cite{conneau-etal-2018-xnli}, KLUE-NLI \cite{park-2021-klue}, and AmericasNLI \cite{ebrahimi-etal-2022-americasnli}), commonsense reasoning (XCOPA \cite{ponti-etal-2020-xcopa} and XStoryCloze \cite{xi-2021-xglm}), anaphora resolution (XWinograd \cite{tikhonov-ryabinin-2021-heads}), and paraphrasing (PAWS-X \cite{yang-etal-2019-paws}).
We perform zero-shot prompting \textit{without any task-specific finetuning} and simply reuse the templates used to prompt the XGLM model \citet{xi-2021-xglm} without performing any prompt engineering.
We translate the prompt templates using automatic translation APIs, and the translated templates can be found in Appendix~\ref{app:prompt-templates}.

%%%%%%%%%%%% baselines %%%%%%%%%%%%%%
\subsection{Baselines}
We compare the adapted BLOOM model against generative multilingual language models which have reported state-of-the-art prompting performance. We also report the prompting performance of the original BLOOM models without any adaptation.

\paragraph{XGLM} XGLM models \cite{xi-2021-xglm} cover 30 natural languages and come in five different numbers of parameters: 564M, 1.7B, 2.9B, 4.9B, and 7.5B. 

\paragraph{mGPT} mGPT \cite{oleh-2022-mgpt} is a GPT model trained on 60 languages from 25 language families using Wikipedia and Colossal Clean Crawled Corpus. It only has 1.3B parameters. 

\paragraph{BLOOMZ and mT0} BLOOMZ and mT0 are BLOOM and mT5 models finetuned on a multilingual task mixture, xP3~\cite{muennighoff2022bloomz}. Here we report performance on the best prompts, which corresponds to instructions being in English while the context and the label are generally non-English. We also do not report performance on PAWS-X data since it is part of the xP3 training mixture.

Among the baselines, XGLM, mGPT, and mT0 have seen all the new languages in Table~\ref{tab:languages} except Guarani during model pretraining.

% \begin{table*}[ht]
% \small
%     \centering
%     \begin{tabular}{llllccc}
%         \toprule
%         Language & Language Family & Word Order & Script & Space-Separated & Seen Script & Contamination \\
%         \midrule
%         German & Indo-European (Germanic) & SVO & Latin & \cmark & \cmark & 0.21\% \\
%         Bulgarian & Indo-European (Slavic) & SVO & Cyrillic & \cmark & \xmark & 0.05\% \\
%         Russian & Indo-European (Slavic) & SVO & Cyrillic & \cmark & \xmark & 0.03\% \\
%         Greek & Indo-European (Hellenic) & SVO & Greek & \cmark & \xmark & 0.03\% \\
%         Turkish & Turkic & SOV & Latin & \cmark & \cmark & 0.03\% \\
%         Korean & Koreanic & SOV & Hangul & \cmark & \xmark & 0.03\% \\
%         Thai & Tai–Kadai & SVO & Thai & \xmark & \xmark & 0.006\% \\
%         Guarani & Tupian & SVO & Latin & \cmark & \cmark & 0\% \\
%         \bottomrule
%     \end{tabular}
%     \caption{Information about the unseen languages used in our experiments. 
%     %We report the contamination of the unseen languages in the 1\% of randomly sampled documents from ROOTS \cite{laurencon-2022-roots,muennighoff2022bloomz}.
%     }
%     \label{tab:languages}
% \end{table*}
\begin{figure*}[ht]
    \includegraphics[width=16cm]{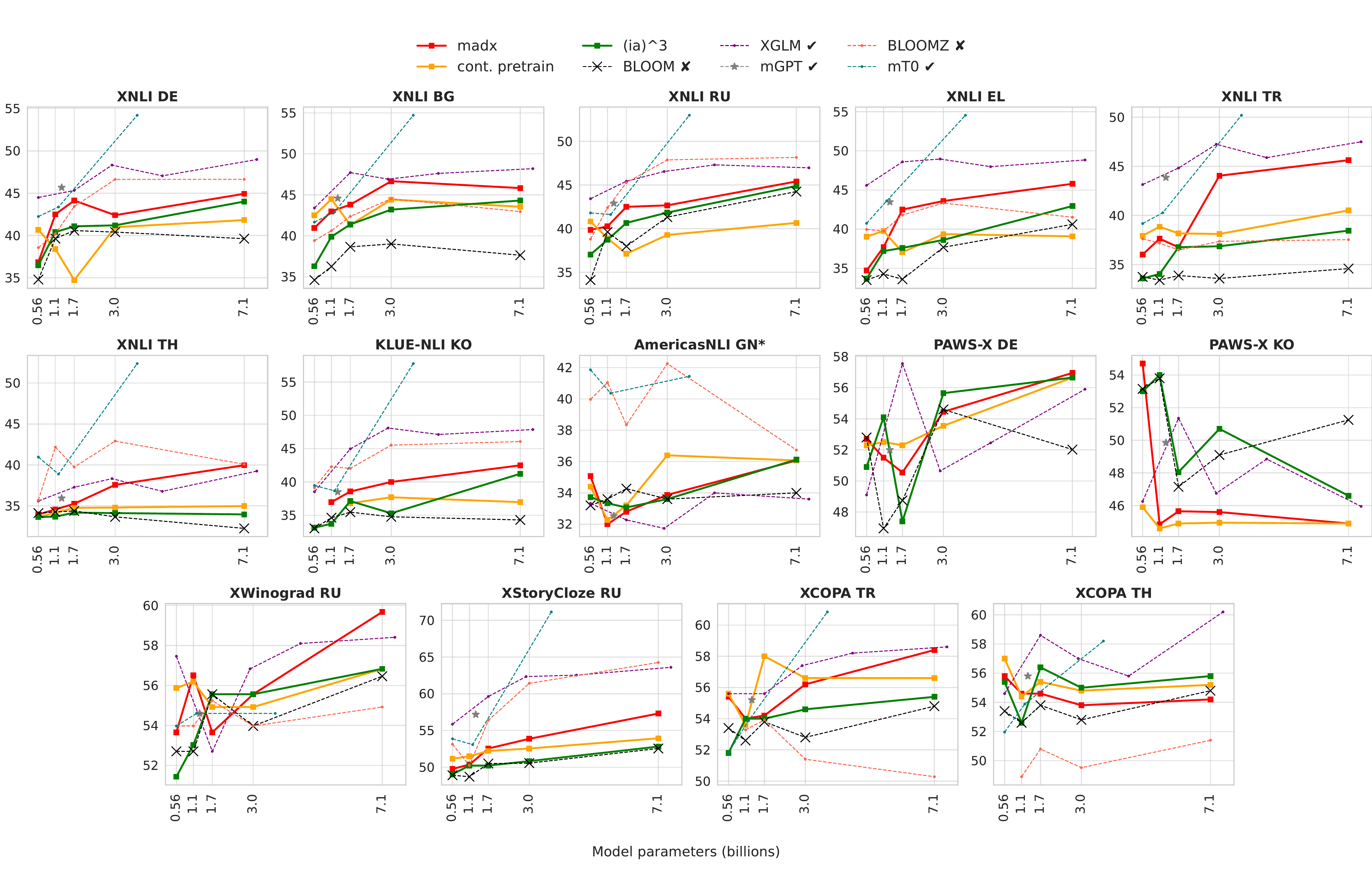}
    \centering
    \caption{Results for zero-shot prompt-based evaluation of natural language inference, commonsense reasoning, anaphora resolution, and paraphrasing tasks. All tasks are evaluated with accuracy measure. Solid lines indicate language adaptation strategies, and dotted lines indicate baselines. $\times$ indicate the non-adapted BLOOM model. Both \cmark~and \xmark~indicate whether the baseline has seen the language during pretraining, except for Guarani (GN) that is unseen for all models. We also ablate BLOOMZ and mT0 from PAWS-X evaluation as the models has been trained on the task.}
    \label{fig:petl}
\end{figure*}

\begin{figure}[ht]
    \includegraphics[width=8cm]{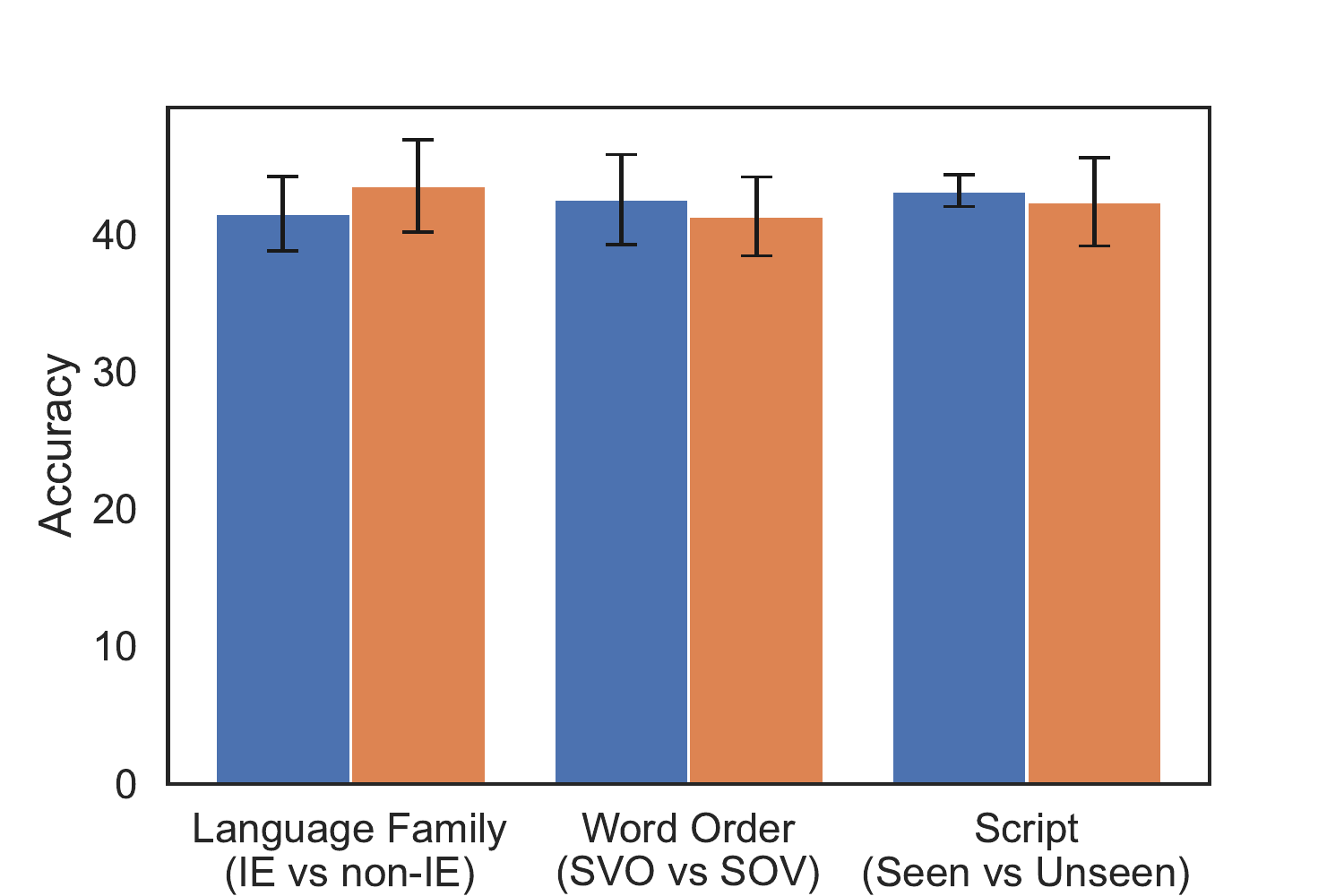}
    \centering
    \caption{Average XNLI prompting performance for different categories of languages, split by whether it belongs to Indo-European (IE) family (left), whether its word order is SVO or SOV (middle), and whether its script system is seen during pretraining (right).}
    \label{fig:language-comparison}
\end{figure}

\begin{figure*}[ht]
    \includegraphics[width=15cm]{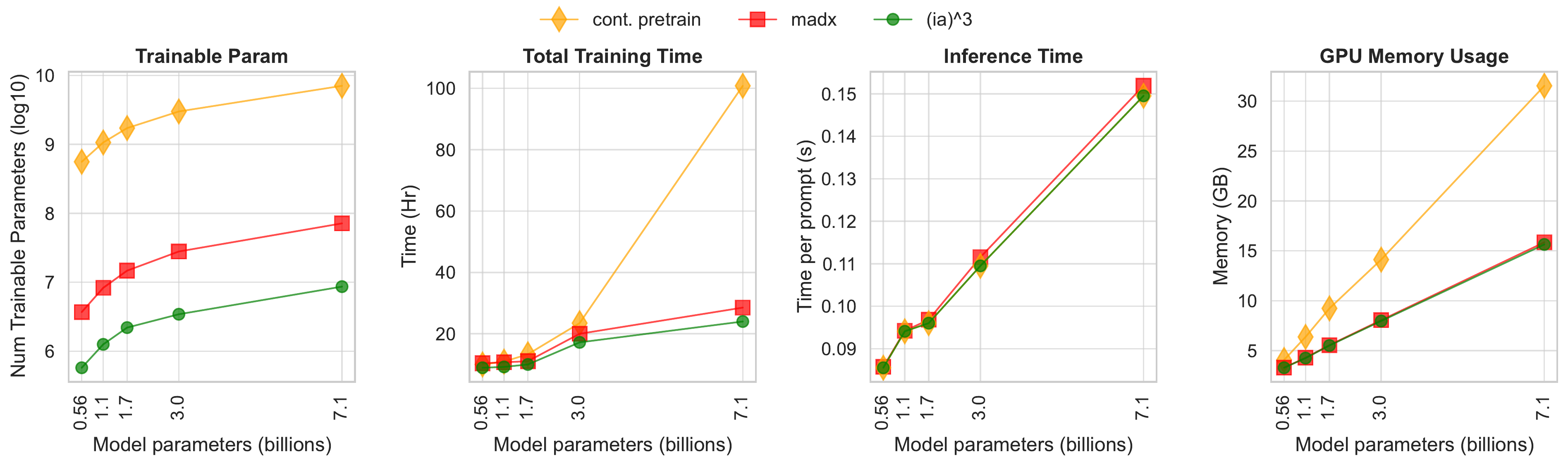}
    \centering
    \caption{Comparison between different language adaptation strategies for BLOOM models on the number of trainable parameters ($\downarrow$), total training time ($\downarrow$), inference `time per prompt on XNLI test set ($\downarrow$), and maximum GPU memory usage ($\downarrow$) on a single A100 GPU machine.}
    \label{fig:system-comparison}
\end{figure*}
%%%%%%%%

%%%% Section
%%%%%%%%
\section{Results and Discussion}
\subsection{Zero-shot Prompting Performance}
\label{sec:result-0-shot-prompting}
Figure~\ref{fig:petl} shows that language adaptation improves the original BLOOM's zero-shot prompting for unseen languages under the resource-constrained setting. Furthermore, in general, language adaptation follows the scaling law which dictates that performance gains correlate with model sizes. We note that when the BLOOM transformer model becomes wider (from 560M to 1.7B parameters), certain tasks such as German XNLI and PAWSX experience performance drops. % We find that on average adapter-based methods obey the scaling law better than continued pretraining.

For the smallest BLOOM model with 560 million parameters, we see that continued pretraining yields the best prompting performance. Our result supports \citeposs{ebrahimi-kann-2021-adapt} findings that continued pretraining of masked language models of similar size, such as mBERT and XLM-Roberta, gives better NER and POS tagging performance than adapters. However, \textbf{when model sizes increases beyond 3 billion parameters, adapter-based language adaptation methods outperform continued pretraining} despite having fewer trainable parameters. 
%This phenomenon has also been observed for English \cite{liu2020tfew} and vision tasks \cite{nayak2022csp}. 
Furthermore, contrary to previous findings \cite{yong2022adapting}, BLOOM adapts well to new languages regardless of their language family, word order, and whether they share the same script system with languages in pretraining data (Figure~\ref{fig:language-comparison}). We note that there are many differences in \citeposs{yong2022adapting} setting. \citet{yong2022adapting} used a multilingual model that uses learned positional embeddings instead of Alibi \cite{press-2022-alibi} and that only supports 13 languages. They also finetuned both the learned positional and word embedding layers. 

We find that the adapted BLOOM matches mGPT's performance in several XNLI tasks and even outperforms XGLM and mT0 on the German PAWS-X and Russian XWinograd tasks. Nonetheless, mT0, which has seen the languages during pretraining and is trained on a multilingual task prompts mixture, exhibits the best zero-shot prompting performance when model parameters are increased.

We find the adapted BLOOM performs poorly on Guarani, which is a truly low-resource language.
Language adaptation only boosts the performance when models beyond 3 billion parameters are used.
We believe this is due to the limited Guarani adaptation training data (30K as opposed to 100K for other languages) as supported by the findings in Section~\ref{sec:lang-adapt-data}.

\paragraph{Best Language Adaptation Strategy} We recommend that the smallest BLOOM model should be adapted with continued pretraining, but larger BLOOM models should be adapted with adapters due to better performance (Figure~\ref{fig:petl}) and compute efficiency (Figure~\ref{fig:system-comparison}). We find MAD-X language adapters give better average zero-shot prompting performance, but (IA)$^3$ adapters have a slight edge in training efficiency due to significantly fewer trainable parameters and smaller training time for larger models. 

\begin{figure}[ht]
\includegraphics[width=7.8cm]{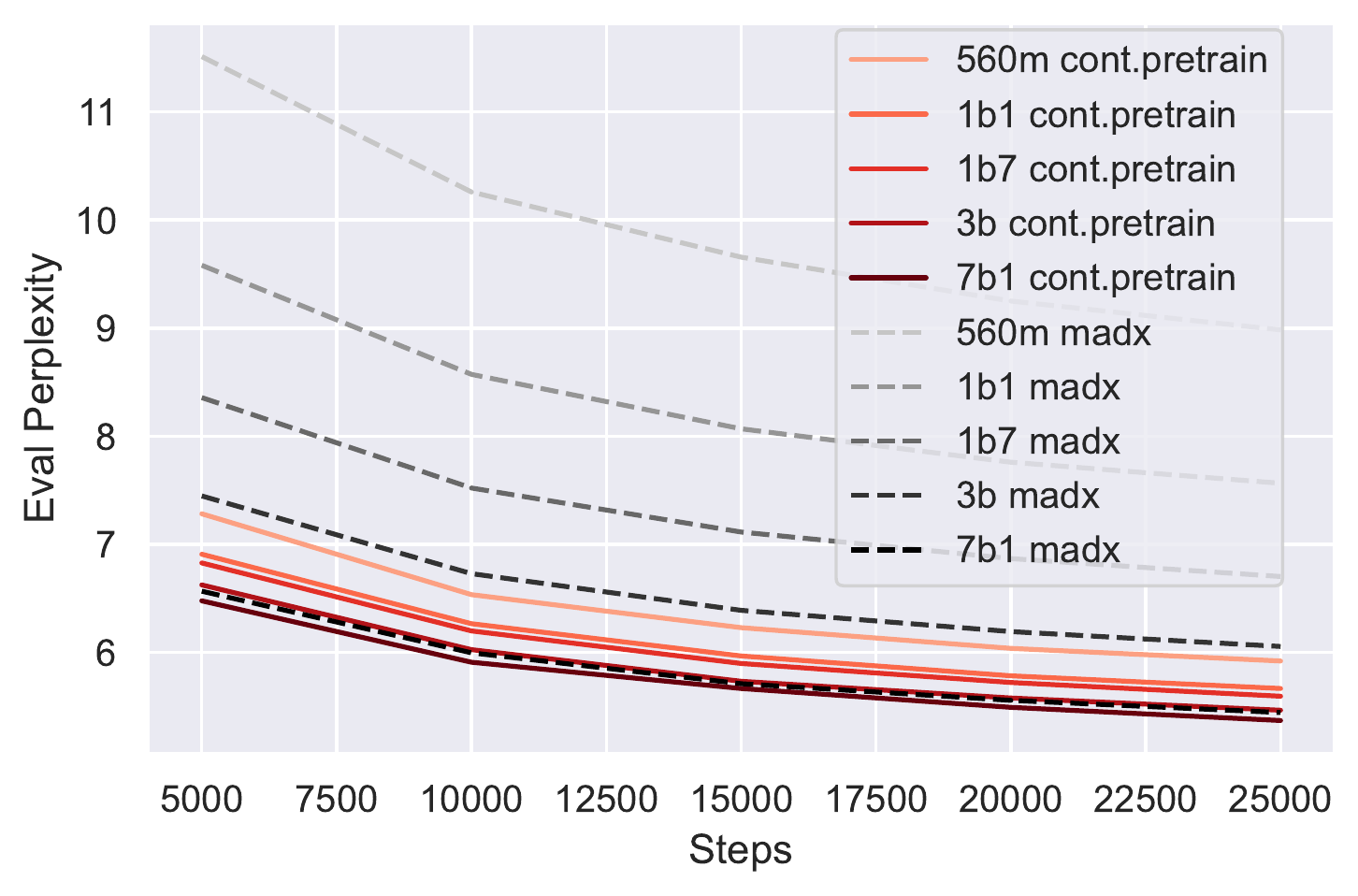}
\centering
\caption{Perplexity curves of continued pretraining and MAD-X language adapters across all BLOOM model sizes on Russian held-out data.}
\label{fig:perplexity-ru}
\end{figure}

\begin{figure*}
    \includegraphics[width=\linewidth]{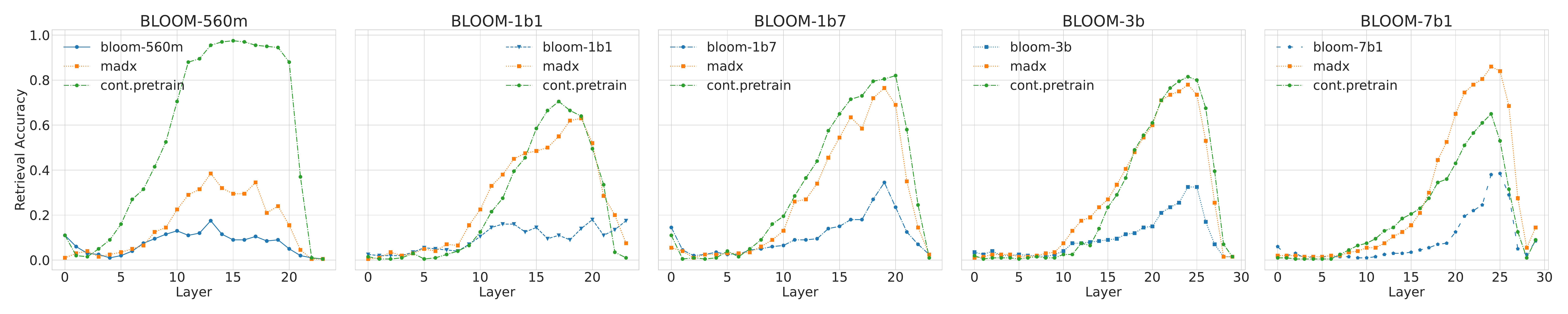}
     \caption{Sentence retrieval accuracy for Russian before and after adaptation with MAD-X adapters and continued pretraining. }
    \label{fig:ru_adapt}   
\end{figure*}

\subsection{Perplexity}
Perplexity can be viewed as a measure of uncertainty when predicting the next token in a sequence, and better language modeling ability means lower perplexity. Figure~\ref{fig:perplexity-ru} shows that evaluation perplexity on Russian texts for continued pretraining and MAD-X language adapters. We find that \textbf{perplexity during language adaptation training does not necessarily correlate with prompting performance.} While perplexity becomes lower for larger models, there is a drop in XWinograd performance for both language adaptation strategies when the model capacity increases from 1.1 billion to 1.7 billion parameters. Furthermore, even though continued pretraining has a lower perplexity than MAD-X language adapters, which suggests that continually-pretrained models better model the Russian OSCAR data, continually-pretrained BLOOM underperform their counterparts for larger model sizes in both XWinograd and XNLI tasks. This finding is in line with \citeposs{LiangEtAl2022} work that highlights the mismatch between perplexity and downstream task performance.

\subsection{Connection to Language Independent Representation}
Figure \ref{fig:ru_adapt} reports sentence retrieval (SR) accuracy for Russian for non-adapted models, as well as models adapted via MAD-X adapters or continued pretraining. We use sentence retrieval accuracy as a way to measure quality of language independent representation, more details in the Appendix~\ref{app:LI}. Note, that in this setting the representations of Russian are based on the adapted model, while representations of English are based on the original model, which excludes the problem of potential catastrophic forgetting. We see that before adaptation, the SR accuracy is very low overall, but bigger model demonstrate better SR results. With adaptation, SR accuracy drastically improves. 

For BLOOM adapted with MAD-X, SR accuracy improves as model grows in parameters. The reason is that adapters' trainable parameters grow in size so they represent Russian sentences better and larger model start from better representations of both languages. \textbf{Interestingly, for continued pretraining, the best SR accuracy result is achieved with the smallest BLOOM model with 560 million parameters}, while larger models achieve much lower SR accuracy. This phenomenon \textit{goes against the scaling law} and is opposite to what has been observed for MAD-X. \footnote{We have observed similar trends for models adapted for German.}

Some previous works \cite{dufter-schutze-2020-identifying} suggest that smaller model would emerge better language-independent representations as it is forced to reuse the same parameters for different languages. However, when model grows it has more freedom to partition its' parameters between languages. Note that this observation has been made in the synthetic settings and to the best of our knowledge has not been confirmed in real multilingual models. Our results in Figure \ref{fig:ru_adapt} could be seen as an additional support to that initial hypothesis. 
When doing continued pretraining with relatively small set of the language adaptation data, there are many ways for the model to optimize it's performance (cf Lottery ticket hypothesis \cite{frankle2018the}). If the model had more freedom to partition its' parameters between different languages, there is no guarantee that the continued pretraining would leverage English-related parameters and therefore could diverge its representation space further away from English.  We hypothesize that this could be a possible explanation of degradation of continued pretraining sentence retrieval accuracy for larger models. 

\subsection{Amount of Language Adaptation Data} \label{sec:lang-adapt-data}
    \begin{figure}[ht]
    \includegraphics[width=8cm]{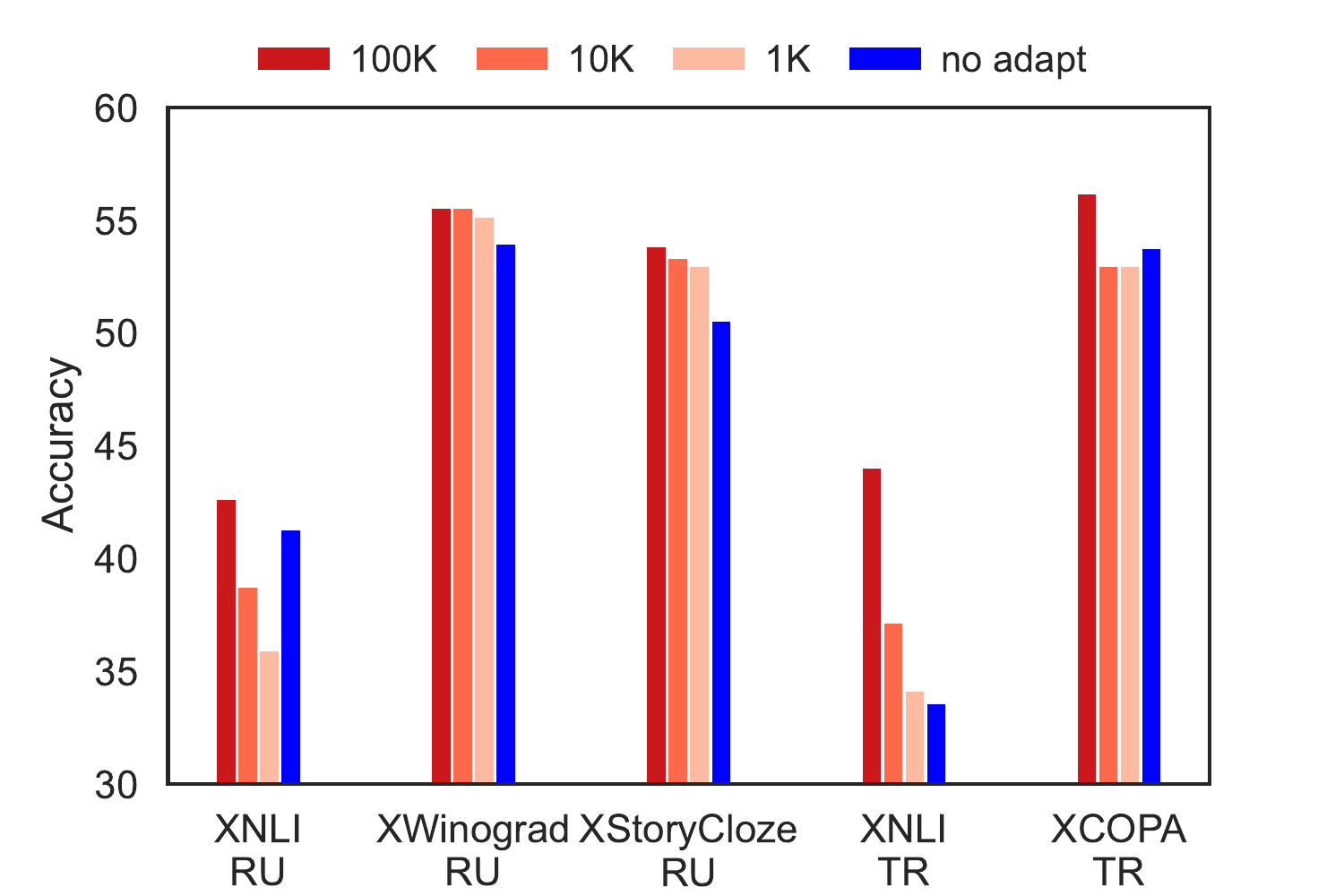}
    \centering
    \caption{Effects of the amount of language adaptation training data on zero-shot prompting of various Russian (RU) and Turkish (TR) tasks. "No adapt" denotes the non-adapted BLOOM model.}
    \label{fig:amount-training-data}
\end{figure}

We simulate different low-resource settings with BLOOM-3B using different amounts of adaptation training data. We use 1K, 10K and 100K samples to simulate different degrees of low-resource settings (see Figure~\ref{fig:oscar-distribution}). Figure~\ref{fig:amount-training-data} demonstrates a positive correlation between the size of adaptation training data and zero-shot prompting performance. We see that, when adapted with less than 100K samples, BLOOM performs worse than its non-adapted counterpart for tasks such as Russian XNLI and Turkish XCOPA. In other words, based on Figure~\ref{fig:amount-training-data} and Table~\ref{tab:100K-num-tokens}, \textbf{we need around 100 million tokens of the new language for effective language adaptation}. However, surprisingly, the extent of the negative effect of low-resource setting can be limited to the type of tasks. For instance, for the same language Russian, we observe a limited effect of low-resource setting on XWinograd and XStoryCloze prompting.

% \subsection{Number of Language Adaptation Steps} 
% Priority 2.

\subsection{Adapters' Capacity}
\begin{figure}[ht]
    \includegraphics[width=8cm]{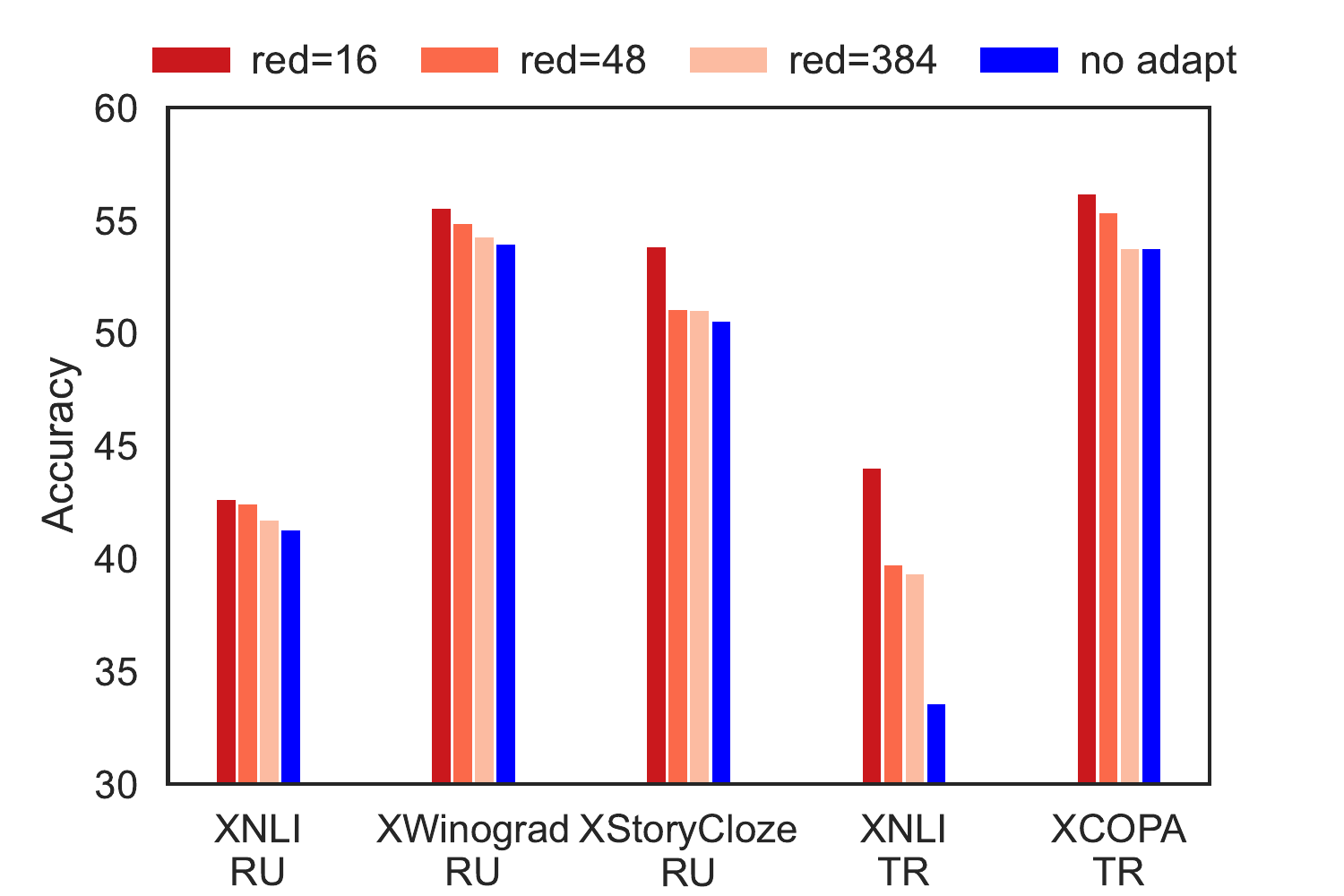}
    \centering
    \caption{Effects of the MAD-X language adapters' reduction factors on zero-shot prompting of various Russian (RU) and Turkish (TR) tasks. "No adapt" denotes the non-adapted BLOOM model.}
    \label{fig:madx-reduction}
\end{figure}

We investigate the effect of the size of adapters' capacity by varying the reduction factor (also known as compression rate \cite{ruckle-etal-2021-adapterdrop}) in the adapter’s bottleneck layer.\footnote{We also investigate the effects of the placement of adapters, invertible adapters, and model pretraining on language adaptation (see Appendix~\ref{app:placement-adapters} and~\ref{app:ablations}).}
A smaller reduction value would lead to a larger amount of adapter parameters. Contrary to \citet{yong2022adapting}, we observe a positive correlation between the amount of adapters' parameters and prompting performance (see Figure~\ref{fig:madx-reduction}).

\subsection{Adapting BLOOMZ}
We also investigate language adaptation strategies for BLOOMZ, which is BLOOM finetuned on many different task prompts to achieve better cross-lingual and cross-task generalization \cite{muennighoff2022bloomz}. 

\begin{figure}[htbp]
    \includegraphics[width=8cm]{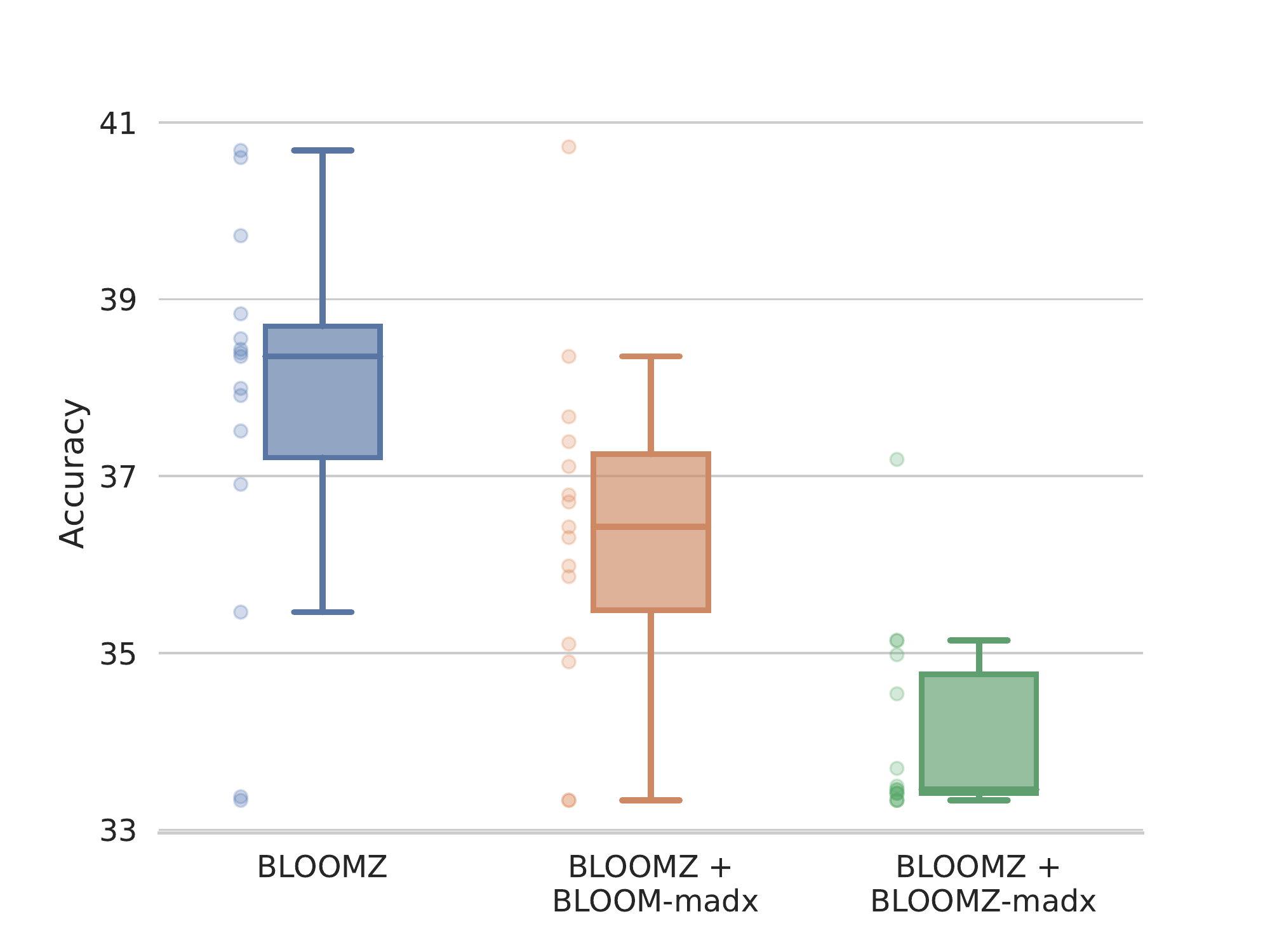}
    \centering  
    \caption{Zero-shot prompting performance of adapted BLOOMZ-560m on German XNLI task. Each dot represents the accuracy of one prompt template, where blue dots indicate the results of non-adapted BLOOMZ and red dots BLOOMZ with adapters.}
    \label{fig:xnli-bloomz-madx}
\end{figure}

\begin{figure*}[htbp]
    \includegraphics[width=15cm]{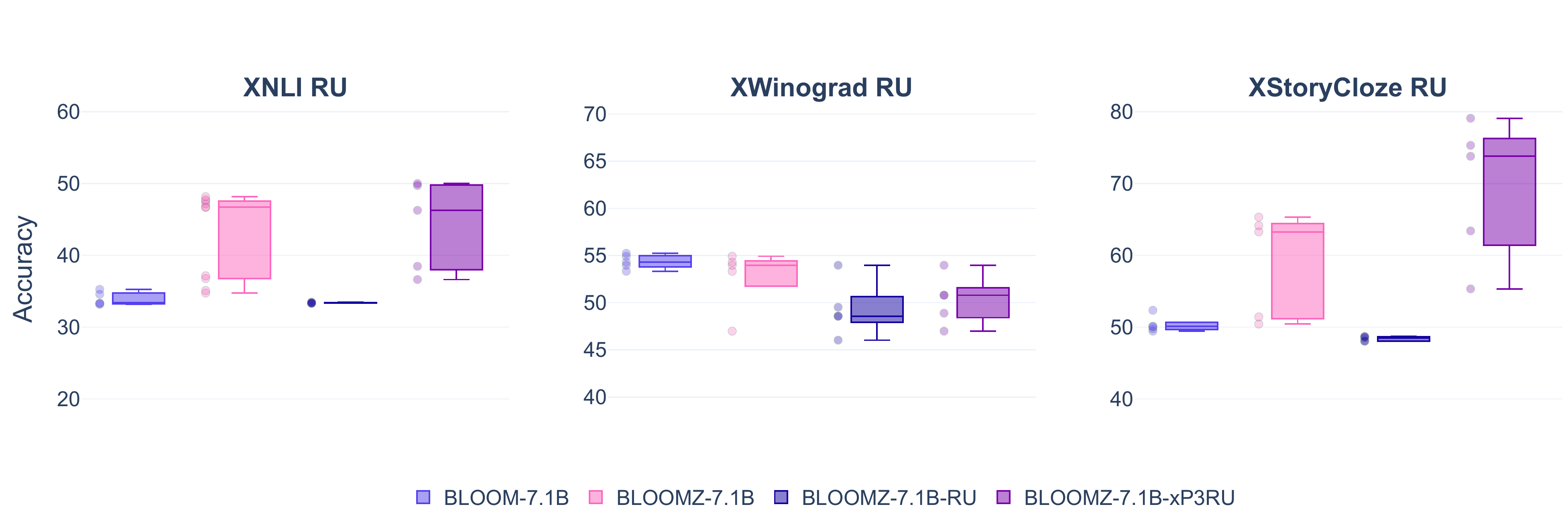}
    \centering
    \caption{Performance on unseen language tasks in Russian of BLOOMZ variants.}
    \label{fig:mtf}
\end{figure*}

\subsubsection{Adding Language Support through Unlabeled Data}

Similar to adapting BLOOM, we train MAD-X language adapters for BLOOMZ using the same experimental setting on monolingual OSCAR data. In Figure~\ref{fig:xnli-bloomz-madx}, we show that BLOOMZ-560m has a median accuracy of around 38.5\% for the German XNLI tasks (left bar), but after language adaptation, it performs the worst with an accuracy as poor as a random classifier at 33\% (right bar). However, when equipped with BLOOM's language adapters (this is possible because BLOOM and BLOOMZ share the same architecture), BLOOMZ retains its prompting ability (middle bar). The result suggests that \textbf{BLOOMZ loses its prompting capability gained from multitask instruction tuning after language adaptation} on the free-form text of monolingual OSCAR corpora.

\subsubsection{Adding Language Support through Instruction Tuning} \label{app:mtf}

We experiment with learning a new language during instruction tuning using the same recipe as BLOOMZ \cite{muennighoff2022bloomz}. We use Russian, which BLOOM models have not intentionally seen during pretraining. We collect supervised natural language task data in Russian and finetune the pretrained 7.1 billion parameter BLOOM model to create two variants: (a) BLOOMZ-7.1B-RU, which is finetuned only on the Russian task data, and (b) BLOOMZ-7.1B-xP3RU, which is finetuned on the full xP3 dataset \cite{muennighoff2022bloomz} with Russian data added to it. We compare the two models with BLOOM-7.1B and BLOOMZ-7.1B in Figure \ref{fig:mtf}.  We find that finetuning on only Russian (BLOOMZ-7.1B-RU) without the other languages and tasks in the xP3 mixture shows only tiny improvements over the pretrained baseline on XStoryCloze. This is likely due to the lack of diversity in the finetuning of BLOOMZ-7.1B-RU \cite{chung2022scaling}, as the Russian-only split contains fewer tasks and prompts than the full xP3 dataset. On the other hand, \textbf{when adding Russian to the instruction tuning mixture (BLOOMZ-7.1B-xP3RU), the performance of the best prompt improves on XNLI and XStoryCloze.} This means that adding new languages during multitask finetuning can be effective but requires additional diverse tasks in other languages. 

% BLOOM-7.1B, which has only gone through pretraining, is at the random baseline of 0.3 on XNLI and 0.5 on XWinograd and XStoryCloze. As shown in \citet{muennighoff2022bloomz}, instruction tu the pretrained model improves XNLI and XStoryCloze Russian performance despite Russian not being part of the finetuning mixture (BLOOMZ-7.1B).

%%% TODO: 
%%%% may add these sections later on after arxiv submission.
% \subsection{Does In-Context Learning Help?}
% \missingfigure{}
% \customtodo{N-shot in-context learning for poor-performing languages.}

% \subsection{Embedding Layers: Frozen, Trainable,

% \subsection{Prompt Designs}
% \subsubsection{Code-Mixed Prompts}
% \customtodo{(Include Results where we use EN prompts.)}

% \subsubsection{P3 Prompts}
% See Figure~\ref{fig:xnli-p3}. Talk about how language adapters are not compatible with P3 prompts, BLOOMZ, and language adaptation causes catastrophic forgetting on P3 prompts.

% \subsection{Catastrophic Forgetting}
% \customtodo{(Include Results to showcase if using original data (50K ROOTS, 50K RU) helps.}
% \missingfigure{}

% \begin{figure}[ht]
% \includegraphics[width=8cm]{assets/xnli-en.pdf}
% \centering
% \caption{Continued pretraining causes catastrophic forgetting on seen languages, regardless of model sizes.}
% \label{fig:xnli-en}
% \end{figure}
%%%%%% Conclusion
\section{Conclusion}
We compare the compute-performance trade-off of different language adaptation strategies for extending BLOOM of various sizes to new languages. Contrary to previous work, we find that adapter-based strategies best adapt larger BLOOM models for prompting under low-resource settings. We also investigate different language adaptation factors such as the size of language adaptation data and capacity of adapters. Finally, we investigate the relationship between language adaptation and instruction tuning using the BLOOMZ model, where we find including new languages during instruction tuning most effective.

% We compare different language adaptation strategies for extending BLOOM of various sizes to new languages, and we find that adapters best adapt large language models under low-resource settings. Our study also shows that including new languages during the instruction tuning of BLOOMZ most effective in adapting instruction-tuned models.

\section{Limitations}
\subsection{Vocabulary and Embedding Adaptation}
We do not explore vocabulary and embedding adaptation. Our models used byte-level tokenization, and therefore can handle unseen scripts. However, one can argue that the tokenization of unseen scripts might be suboptimal. For instance, languages with unseen script will require longer post-tokenization, therefore impacting the performance efficiency.
\citet{koto2021indobertweet} have shown that when adapting to a new domain, LM achieved better performance, despite the fact that the old vocabulary can support the new domain as well. Exploring the quality impact of token adaptation for new languages and new scripts would be very interesting.
In parallel, exploring the best way to initialize embeddings of the newly formed tokens is also interesting.

\subsection{Parameter-Efficient Finetuning Strategies}
We have only considered a limited number of parameter-efficient finetuning strategies (see Section~\ref{sec:lang-adapt-strats} and Appendix~\ref{app:lang-adapt-strats}) due to computational constraints. Nonetheless, we believe that other strategies such as prompt tuning \cite{lester-etal-2021-power,tu2022prompt} and ladder side-tuning \cite{sung2022lst} can adapt BLOOM as well as the adapter-based strategies explored in our experimental setting. Recent work has also shown that combining different types of parameter-efficient finetuning methods, including adapters, can lead to better performance \cite{mao-etal-2022-unipelt,he2022mixandmatch}. As we recommend adapter-based language adaptation for larger language models, it would be interesting to explore methods that combine adapters for better prompting performance.

\subsection{Low-Resource Languages}
One limitation of our work is that our set of new languages only covers one truly low-resource language, which is Guarani. As our work shows that 100 million tokens are needed for effective adaptation to prompt in a new language (see Section~\ref{sec:lang-adapt-data}), a truly low-resource language usually lacks sufficient unlabeled data for such adaptation \cite{joshi-etal-2020-state}. Therefore, we urge the community to study data-efficient methods for adapting large language models to prompt under an extremely low-resource setting.

\subsection{Generative Tasks}
Since we only cover natural language understanding tasks in our experimental setup, our findings may not generalize to generation tasks such as summarization. Furthermore, language adaptation on monolingual data can lead to catastrophic forgetting of seen languages (see Appendix~\ref{app:catastrophic-forgetting}); therefore, adapted models are not suitable for multilingual generative tasks that require an understanding of multiple languages such as machine translation. Future work is needed for studying solutions to mitigate catastrophic forgetting. 

\subsection{Experimental Settings}
We used the sequence length of 1024 by mistake (instead of 2048 as described in \citet{scao-2022-bloom}) as we followed prior work on adapting BLOOM models to new languages \cite{yong2022adapting}. However, in principle, it should not change the conclusions we draw from our study since none of the evaluation tasks are done on sequences longer than 1024 tokens. Our post-hoc experimental results with the correct sequence length of 2048 (see Appendix~\ref{app:post-hoc}) also align with our results discussed in Section~\ref{sec:result-0-shot-prompting}.

We did not carry out adaptation for the largest BLOOM model and BLOOMZ model with 176 billion parameters due to prohibitive computational costs. We leave them for future work to explore language adaptation for language models with hundreds of billions of parameters.

% \section{Ethics Statement and Broader Impact}
% \input{sections/acknowledgments}

\bibliographystyle{acl_natbib}
\bibliography{main,anthology}

\appendix
\section*{Appendix}

\section{Authors' Contributions}
Our work extended the language support of the BLOOM model \cite{scao-2022-bloom} that was created under the BigScience project, a year-long initiative to create open-source large multilingual language models in a transparent manner which involves 600 researchers from over 50 countries and 250 institutions. All authors came from the BigScience multilingual modeling working group, and in the following list, we document our contributions made to this work. 

\noindent
\textbf{Zheng-Xin Yong} led the project, set up training and evaluation pipelines, coordinated resources and experiments, and wrote most of the paper.

\noindent
\textbf{Vassilina Nikoulina} advised the project.

\noindent
\textbf{Zheng-Xin Yong and Vassilina Nikoulina} initially conceptualized the project. 

\noindent
\textbf{Zheng-Xin Yong, Hailey Schoelkopf, and Lintang Sutawika} implemented various parameter-efficient finetuning methods. 

\noindent
\textbf{Zheng-Xin Yong, Hailey Schoelkopf, Alham Fikri Aji, David Ifeoluwa Adelani, Khalid Almubarak, M Saiful Bari, Ahmed Baruwa, Jungo Kasai, and Vassilina Nikoulina} performed language adaptation training and prompting evaluation to collect results.

\noindent
\textbf{Zheng-Xin Yong and Niklas Muennighoff} performed BLOOMZ language adaptation experiments.

\noindent
\textbf{Vassilina Nikoulina} performed the sentence retrieval experiments.

\noindent
\textbf{Zheng-Xin Yong, Hailey Schoelkopf, Niklas Muennighoff, Alham Fikri Aji, David Ifeoluwa Adelani, Khalid Almubarak, Ahmed Baruwa, Jungo Kasai, Genta Indra Winata, Stella Biderman, Edward Raff, Dragomir Radev, and Vassilina Nikoulina} contributed to the paper.

%%%%%%%%%%%%%%%%%%%%%%%%%%%%%%%%%%%%%%%%%%%%%%
%%%%%%%%%%%%%%%%%%%%%%%%%%%%%%%%%%%%%%%%%%%%%%

\begin{figure*}

    \centering
    \includegraphics[width=\linewidth]{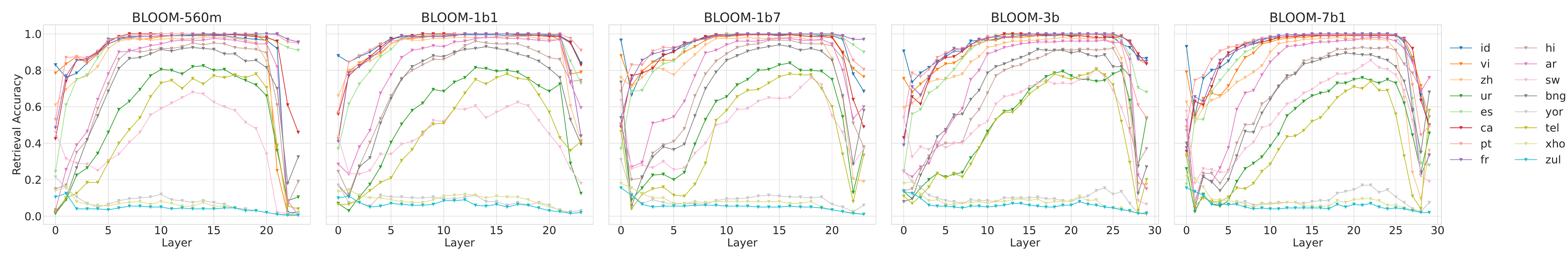}
    \caption{Sentence Retrieval accuracy for known languages for different BLOOM models across layers.}
    \label{fig:known_li}   
\end{figure*}

\section{How does Language Independent Representation changes with Model Sizes}\label{app:LI}
In this work we try to establish the connection between the quality of \textit{language-independent} representation a pretrained LM can emerge, and its adaptability to the new language. In order to evaluate the quality of \textit{language-independent} representation we rely on sentence retrieval task (similar to \cite{dufter-schutze-2020-identifying, artetxe-schwenk-2019}) computed on FLORES dataset\footnote{We take a subset of 200 sentences of the dev set}. Sentence retrieval task is to identify closest sentence in English given a representation of the sentence in the new language, which imitates most most popular knowledge transfer scenario, where we have final task data available in English only. 
In addition to what has been done previously, we compute sentence retrieval accuracy at each layer of the different pretrained models, to better understand where and how the language-independent represetnation emerges. 
Figure \ref{fig:known_li} reports the sentence retrieval accuracy for the subset of languages used to train BLOOM model, for different model sizes. We notice that all the models follow very similar pattern: first and last layers of the model show quite low SR accuracy, but intermediate layers are able to achieve almost perfect sentence retrieval accuracy for all model sizes. An exception is a set of very low-resource languages which seem to have very low Sentence Retrieval Accuracy from English. We do not notice any significant between models of different sizes for the languages that have been observed during training.

\section{Batch Sizes}
\begin{figure}[ht]
\includegraphics[width=7.5cm]{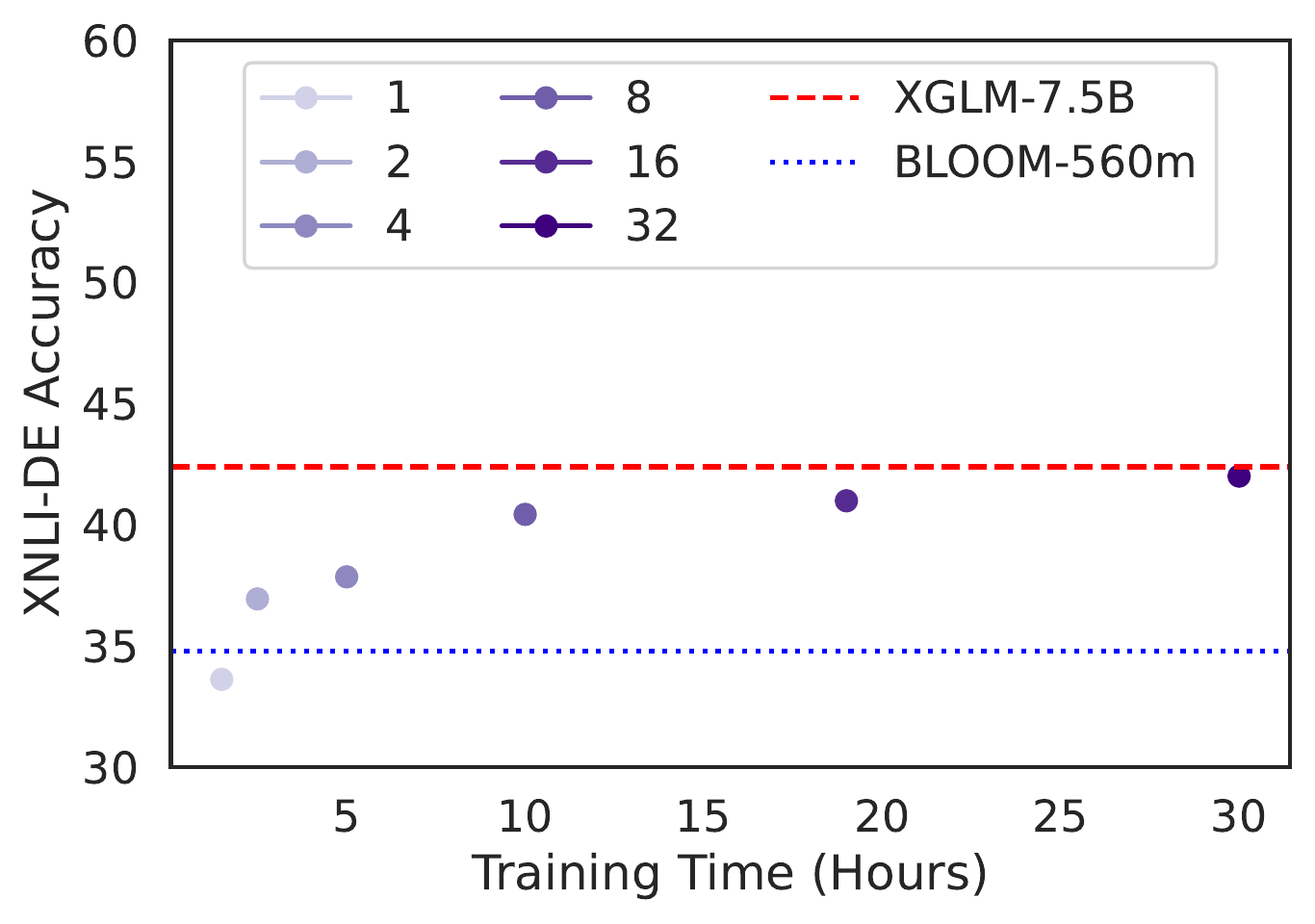}
\centering
\caption{German XNLI prompting performance with the BLOOM-560m model trained with various batch sizes of monolingual language adaptation data.}
\label{fig:xnli-de-bsz}
\end{figure}

Figure~\ref{fig:xnli-de-bsz} shows that the batch size of 8 is an optimal batch size considering the performance-compute trade-off. Performance increases quickly when batch size increases to 8 and slowly afterward.

\begin{figure}[ht]
\includegraphics[width=8cm]{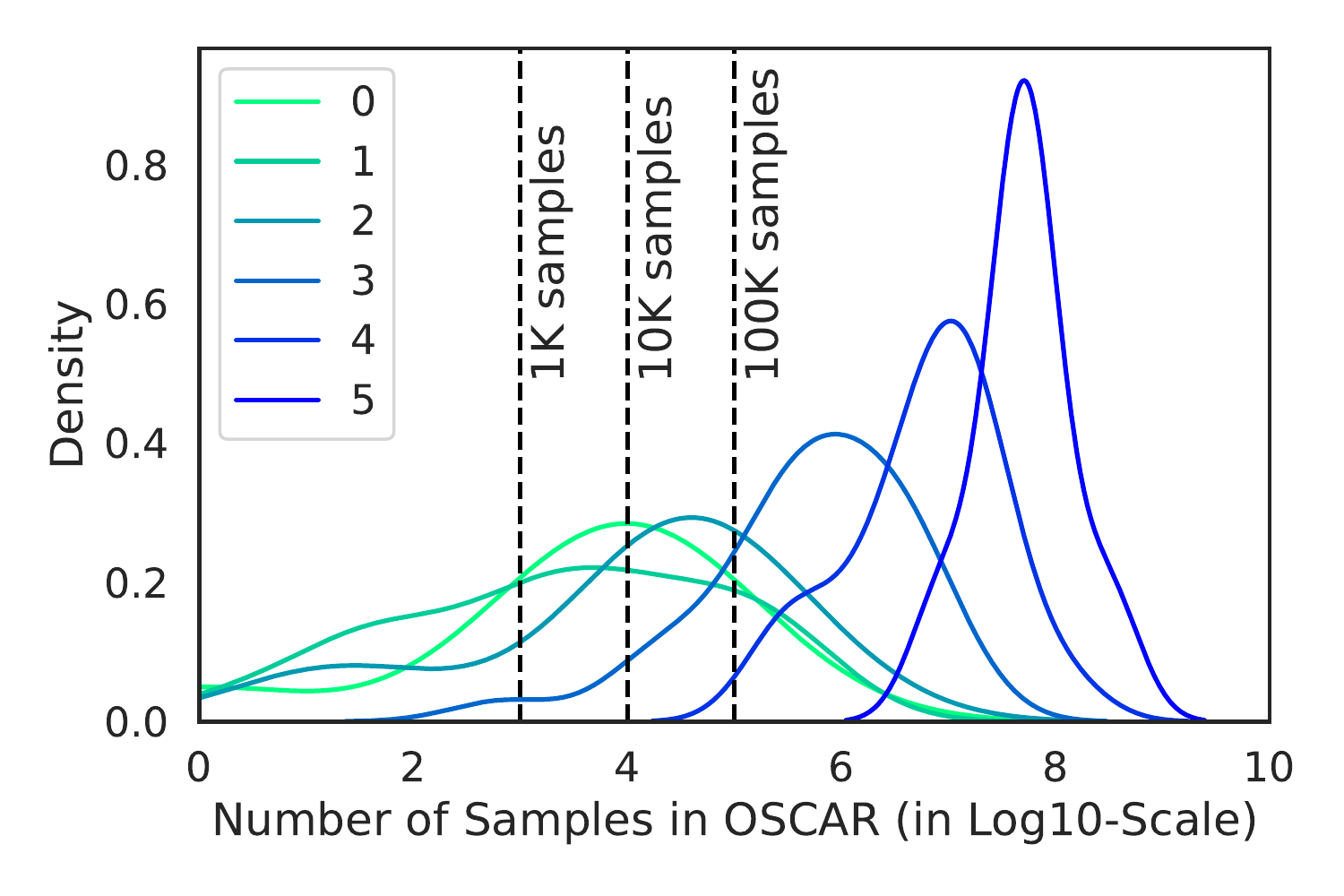}
\centering
\caption{Distribution of language resources on OSCAR \cite{ortiz:oscar_2019} grouped by the level of resource setting (0 indicates very low-resource, 5 indicates high-resource) according to \citet{joshi-etal-2020-state}.}
\label{fig:oscar-distribution}
\end{figure}

%%% commented out because it's an isolated figure. 
% \begin{figure}[ht]
% \includegraphics[width=7.5cm]{assets/xnli-de-training-steps.pdf}
% \centering
% \caption{Longer continued pretraining steps do not necessarily mean better downstream task performance. X mark is used to indicate the smallest validation loss during language adaptation training, and we see that the lowest validation loss does not necessarily indicate the best prompting performance.}
% \label{fig:xnli-de-training-steps}
% \end{figure}

\section{Composable Sparse-Finetuning} \label{app:c-sft}
Composable Sparse-Finetuning (C-SFT) is a sparse-finetuning method that finetunes language-specific and task-specific sparse subset of language model's parameters (mask), both of which demonstrates composability \cite{ansell-etal-2022-composable}. Since the authors demonstrate that this method outperforms MAD-X in language adaptation for POS and NER tasks, we also experimented with it on prompting. In our setting, we only finetuned the language-specific mask, and we followed \citet{ansell-etal-2022-composable} by freezing the output embedding and all layer normalization parameters. We reused the same hyperparameters but with an even split of 12,500 steps in both first and second stage of C-SFT. We ran our experiments using the publicly released code \url{https://github.com/cambridgeltl/composable-sft/tree/6e3ef08cf0fc465d59285e529569387246028538}. 

\begin{figure}
    \includegraphics[width=\linewidth]{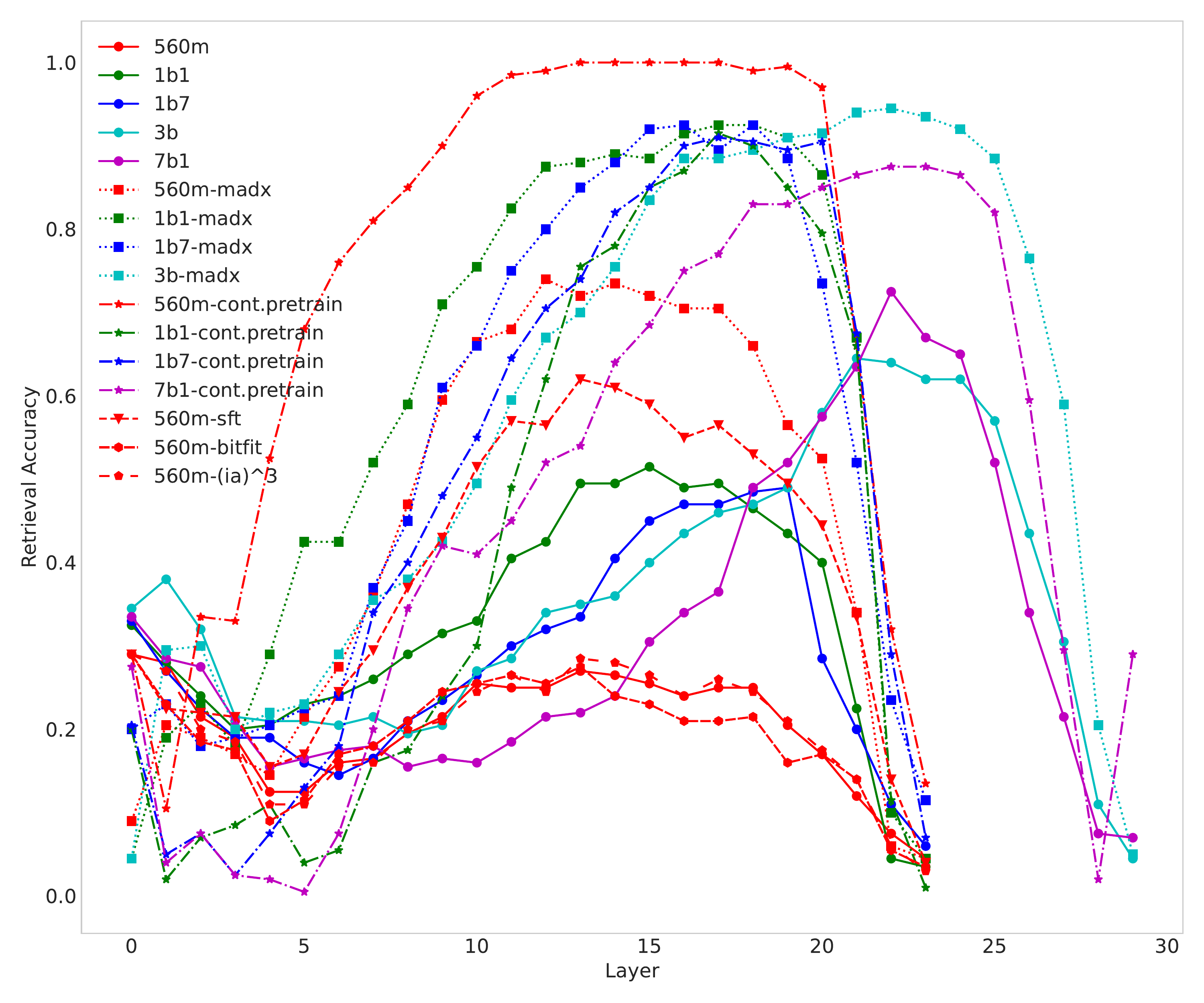}
     \caption{Sentence retrieval accuracy for German with different language adaptation strategies. }
    \label{fig:sr-all-lang-adapt}   
\end{figure}

Our preliminary results with smaller BLOOM models show that models adapted by C-SFT are not capable of prompting (see Table~\ref{tab:composable-sft}) even though it improves sentence retrieval score (red $\blacktriangledown$ in Figure~\ref{fig:sr-all-lang-adapt}). In addition to the poor prompting performance, C-SFT requires finetuning the entire model and needs twice the GPU RAM memory than continued pretraining for storing a copy of the original model to compute the sparse mask. We found that we can improve prompting performance with longer C-SFT training steps. When we ran 25K training steps for both stages of C-SFT, totalling 50K language adaptation steps (instead of 25K total steps), German XNLI prompting performance improved from 33.01\% to 35.97\%. However, due to computational constraint, we did not run more experiments with C-SFT.

% It also lacks Deepspeed integration in the current version of the released code, which is necessary for adapting large BLOOM models. 

\begin{table}[ht]
\small
    \centering
    \begin{tabular}{lcccc}
        \toprule
        Models & Adapt. & DE & RU & TR \\
        \midrule
        Random & - & 33.33\% & 33.33\% & 33.33\%\\
        \midrule
        BLOOM-560m & - & 34.79\% & 34.11\% & 33.75\% \\
        BLOOM-560m & MAD-X & 36.83\% & 39.86\% & 36.03\%\\
        BLOOM-560m & C-SFT & 33.01\% & 	33.05\% & 33.39\%\\
        \midrule
        BLOOM-1b1 & - & 39.64\% & 39.62\% & 33.43\% \\
        BLOOM-1b1 & MAD-X & 42.5\% & 40.26\% & 37.64\%\\
        BLOOM-1b1 & C-SFT & 34.93\% & 33.49\% & 33.39\%\\
        \bottomrule
    \end{tabular}
    \caption{XNLI Accuracy for unadapted BLOOM model, MAD-X language adapters, and Composable Sparse-Finetuning (C-SFT).}
    \label{tab:composable-sft}
\end{table}

\section{Korean PAWS-X}
\begin{figure}
    \centering
    \includegraphics[width=\linewidth]{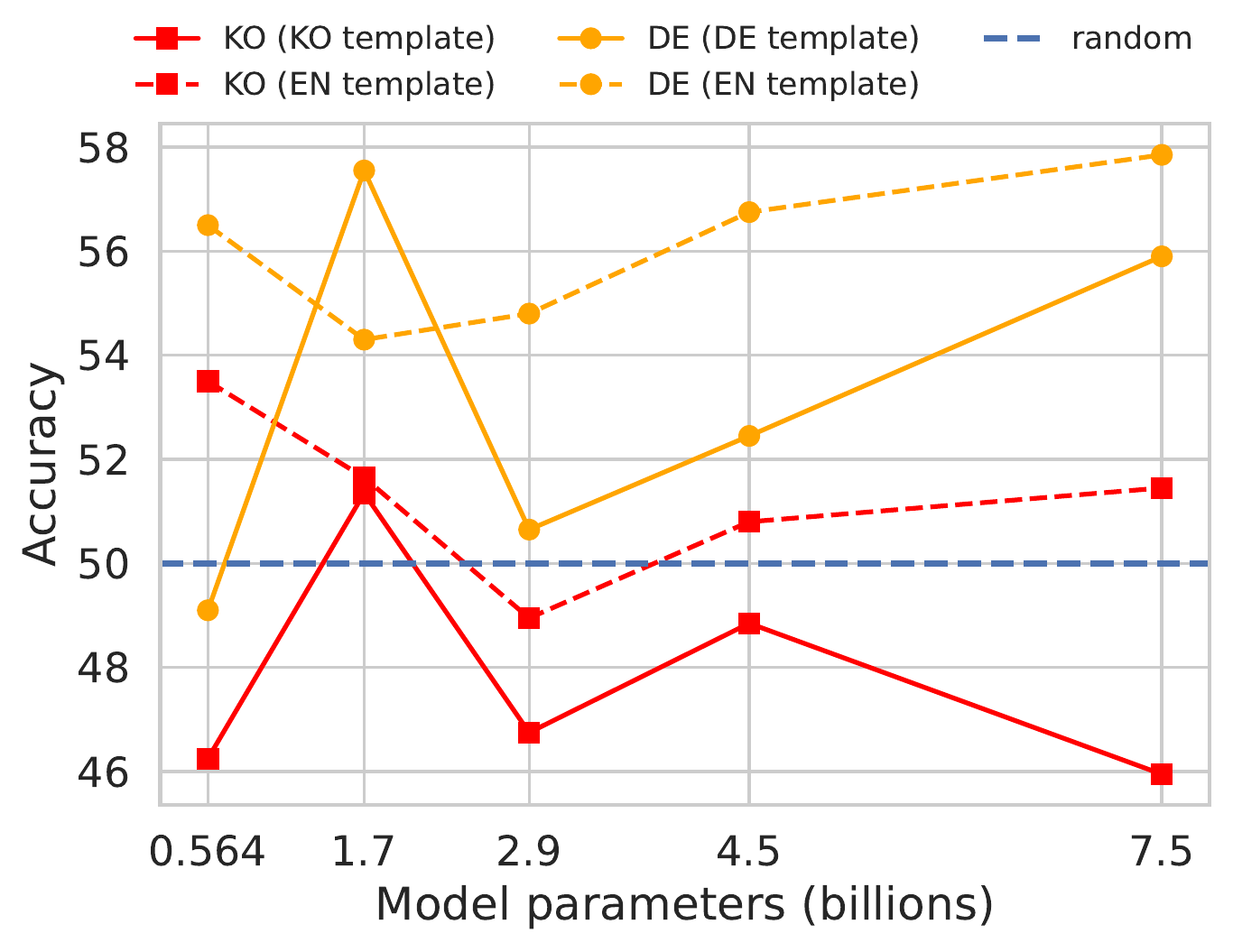}
    \caption{XGLM model's zero-shot prompting performance on German and Korean PAWS-X task with prompt templates in its own language or English language.}
    \label{fig:ko_pawsx}   
\end{figure}

Figure~\ref{fig:petl} shows that all models perform poorly on the Korean PAWS-X task, where a random classifer baseline scores 50\%. Our analysis with English templates shows that XGLM baseline, which is effective at code-mixed prompting setting \cite{xi-2021-xglm}, also performs poorly for Korean PAWS-X (see Figure~\ref{fig:ko_pawsx}). Therefore, we believe that the prompt template is ineffective for Korean PAWS-X task. 

\section{Prompt Templates} \label{app:prompt-templates}
We used the same templates proposed by \cite{xi-2021-xglm} for prompting the XGLM model. Table~\ref{tab:prompt-templates} shows the English and translated templates for all the tasks. We did not manage to get Thai templates rendered with pdflatex, but the templates can be found on \href{https://github.com/yongzx/lm-evaluation-harness/blob/xglm-prompt/lm_eval/tasks/xnli.py}{here for XNLI} and \href{https://github.com/yongzx/lm-evaluation-harness/blob/xglm-prompt/lm_eval/tasks/xcopa.py}{here for XCOPA}.

\begin{table*}[ht]
\small
    \centering
    \begin{tabular}{lccc}
        \toprule
        Tasks & Languages & Templates & Verbalizers \\
        \midrule
        \multirow{6}{*}{XNLI} & EN & \textsc{\{Premise\}}, right? [Label], \textsc{\{Hypothesis\}} & Yes $\vert$ No $\vert$ Also \\
         & BG & \textsc{\{Premise\}}, нали? [Label], \textsc{\{Hypothesis\}} & Да $\vert$ Не $\vert$ Освен това \\
         & DE & \textsc{\{Premise\}}, richtig? [Label], \textsc{\{Hypothesis\}} & Ja $\vert$ Nein $\vert$ Auch \\
         & EL & \textsc{\{Premise\}}, σωστά; [Label], \textsc{\{Hypothesis\}} & Ναι $\vert$ Όχι $\vert$ Επίσης\\
         & RU & \textsc{\{Premise\}}, не так ли? [Label], \textsc{\{Hypothesis\}} & Да $\vert$ Нет $\vert$ А также \\
         % & TH & \customtodo{Here} \\
         % & TH & \textsc{\{Premise\}} ใช่ไหม [Label], \textsc{\{Hypothesis\}} & ใช่ $\vert$ ไม่ $\vert$ นอกจากนี้ \\

         \midrule
         
         KLUE-NLI & KO & \textsc{\{Premise\}}, 맞지? [Label], \textsc{\{Hypothesis\}} & 예 $\vert$ 아니요 $\vert$ 또한 \\

         \midrule
         
         AmericasNLI & GN & \textsc{\{Premise\}}, ¿ajépa? [Label], \textsc{\{Hypothesis\}} &  He\~e $\vert$ Nah\'aniri $\vert$ Ave \\
        
        \midrule
        \multirow{3}{*}{PAWS-X} & EN & \textsc{\{Sentence 1\}}, right? [Label], \textsc{\{Sentence 2\}} & Yes $\vert$ No\\
         & DE & \textsc{\{Sentence 1\}}, richtig? [Label], \textsc{\{Sentence 2\}} & Ja $\vert$ Nein \\
         & KO & \textsc{\{Sentence 1\}}, 맞죠? [Label], \textsc{\{Sentence 2\}} & 예 $\vert$ 아니오 \\

        \midrule
        \multirow{2}{*}{XStoryCloze} & EN & \multirow{2}{*}{\textsc{\{Context\}} [Label]}  & \multirow{2}{*}{Identity} \\
        & RU & & \\

        \midrule
        \multirow{2}{*}{XWinograd} & EN & \multirow{2}{*}{\textsc{\{Context\}}  \textit{(with '\_' replaced by \normalfont{[Label]})}}  & \multirow{2}{*}{Identity} \\
         & RU & & \\

        \midrule
        \multirow{4}{*}{XCOPA} & \multirow{2}{*}{EN} & \textit{cause:} \textsc{\{Sentence 1\}} because [Label]  & \multirow{4}{*}{Identity} \\
        & & \textit{effect:} \textsc{\{Sentence 1\}} so [Label]\\
        & \multirow{2}{*}{TR} & \textit{cause:} \textsc{\{Sentence 1\}} çünkü [Label] & \\
        & & \textit{effect:} \textsc{\{Sentence 1\}} yani [Label] \\
        % & \multirow{2}{*}{TH} & \customtodo{here} \\
        
        \bottomrule
    \end{tabular}
    \caption{Task templates for prompting BLOOM where "[Label]" is replaced with the answer choices in the verbalizers column. *NLI tasks' verbalizers correspond to entailment, contradiction, and neutral respectively, and PAWS-X's corresponds to true and false respectively. Identity verbalizer maps candidate choice to itself in multiple-choice tasks. }
    \label{tab:prompt-templates}
\end{table*}

\section{Other Parameter-Efficient Finetuning Strategies}
\label{app:lang-adapt-strats}
We experimented with various parameter-efficient finetuning strategies for language adaptation, including BitFit \cite{ben-zaken-etal-2022-bitfit}, (IA)$^3$ \cite{liu2020tfew}, LoRA \cite{hu2022lora}, and FishMask \cite{guo-etal-2021-parameter}. We reported the best result from the two sets of hyperparameters, one reported in the original papers proposing the methods and the other reported in Appendix~\ref{app:lang-adapt-details}). On German XNLI task, we found that MAD-X language adapters still yield the best prompting performance (see Table~\ref{tab:lang-adapt-strats}).

\begin{table}[ht]
\small
    \centering
    \begin{tabular}{lc}
        \toprule
        Adapt. & Accuracy \\
        \midrule
        No Adaptation & 34.79 \\
        \midrule
        MAD-X (Bottleneck adapters) & \textbf{36.83} \\
        BitFit & 33.95 \\
        (IA)$^3$ & 36.31 \\
        (IA)$^3$ + invertible adapters & 36.47 \\
        LoRA & 35.79 \\
        FishMask & 35.59 \\
        \bottomrule
    \end{tabular}
    \caption{German XNLI prompting performance with the BLOOM-560m model adapted by various parameter-efficient finetuning methods.}
    \label{tab:lang-adapt-strats}
\end{table}

\section{Language Adaptation Experimental Setup Details} \label{app:lang-adapt-details}
We trained for a total of 25,000 steps with a batch size of 8 and sequence length of 1024 on the monolingual corpora of the new language. In other words, the models are trained on around 204 million tokens. We evaluated every 5,000 steps on the perplexity of 1,000 held-out validation samples, and we took the best checkpoint for downstream prompting tasks. We defaulted to using a single RTX 3090 GPU machine for each language adaptation training, unless the model is too large or takes too long to run (for instance, performing continued pretraining for BLOOM with 7.1 billion parameters), which we would use eight A100 GPUs with 40GB RAM for training. We conducted single runs for each language adaptation due to computational constraint. 

We performed hyperparameter search on learning rates of \{1e-3, 1e-4, 1e-5\}, linear and cosine decay, and warm-up ratio of \{0, 0.05, 0.1\} using the Russian XNLI task and BLOOM-560m and -1b1 models. Table~\ref{tab:lang-adapt-details} reports the best set of hyperparameters. In general, we found that different sets of hyperparameters caused around 1$\sim$2 \% small difference in XNLI accuracy. Since our primary goal was to study trends and performance-compute trade-offs for language adaptation strategies, we did not perform extensive hyperparameter search.

\begin{table}[ht]
\small
    \centering
    \begin{tabular}{lccc}
        \toprule
        Adapt. & LR & Decay & Warm-up Ratio \\
        \midrule
        Continued Pretraining & 1e-4 & Linear & 0 \\
        MAD-X & 1e-4 & Linear & 0 \\
        (IA)$^3$ & 1e-4 & Linear & 0.1 \\
        \bottomrule
    \end{tabular}
    \caption{Best set of hyperparameters for language adaptation strategies.}
    \label{tab:lang-adapt-details}
\end{table}

\section{Number of Tokens for Language Adaptation Data} \label{app:100K-num-tokens}

We report the number of tokens after preprocessed by BLOOM's BPE tokenizer for all the language adaptation training samples in Table~\ref{tab:100K-num-tokens}. 

\section{Placement of Adapters}\label{app:placement-adapters}
\begin{figure}
    \centering
    \includegraphics[width=0.9\linewidth]{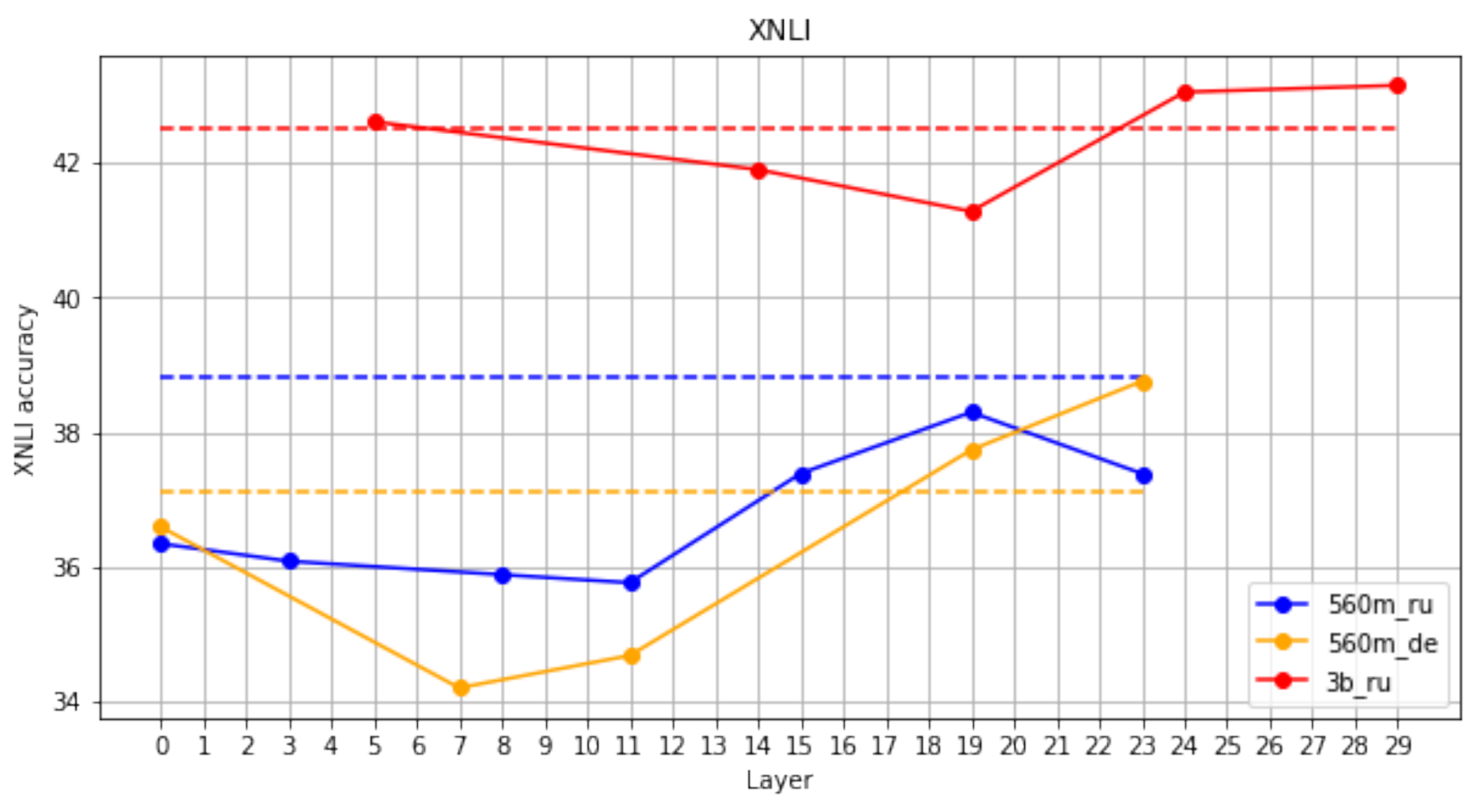}
    \caption{Impact of adapter placement on the quality of model adaptation. Dashed line corresponds to the adapted model with an adapter placed at each layer (referred as mad-x in other experiments). }
    \label{fig:adapt_place_xnli}
\end{figure}

We examined how adapters' placement impacts the overall performance. For this, we kept a single adapter at different layers of the model, where we increased the bottleneck size in a way to match the same parameter count of the model with a full set of adapters\footnote{For model with 24 layers it would result into 24x larger bottleneck size of the adapter.}. Figure \ref{fig:adapt_place_xnli} compares adapter placement results on XNLI task. We note that layers in the middle benefit less from the language adaptation, and the last layers benefit most from the language adaptation.

\begin{figure}[ht]
\includegraphics[width=8cm]{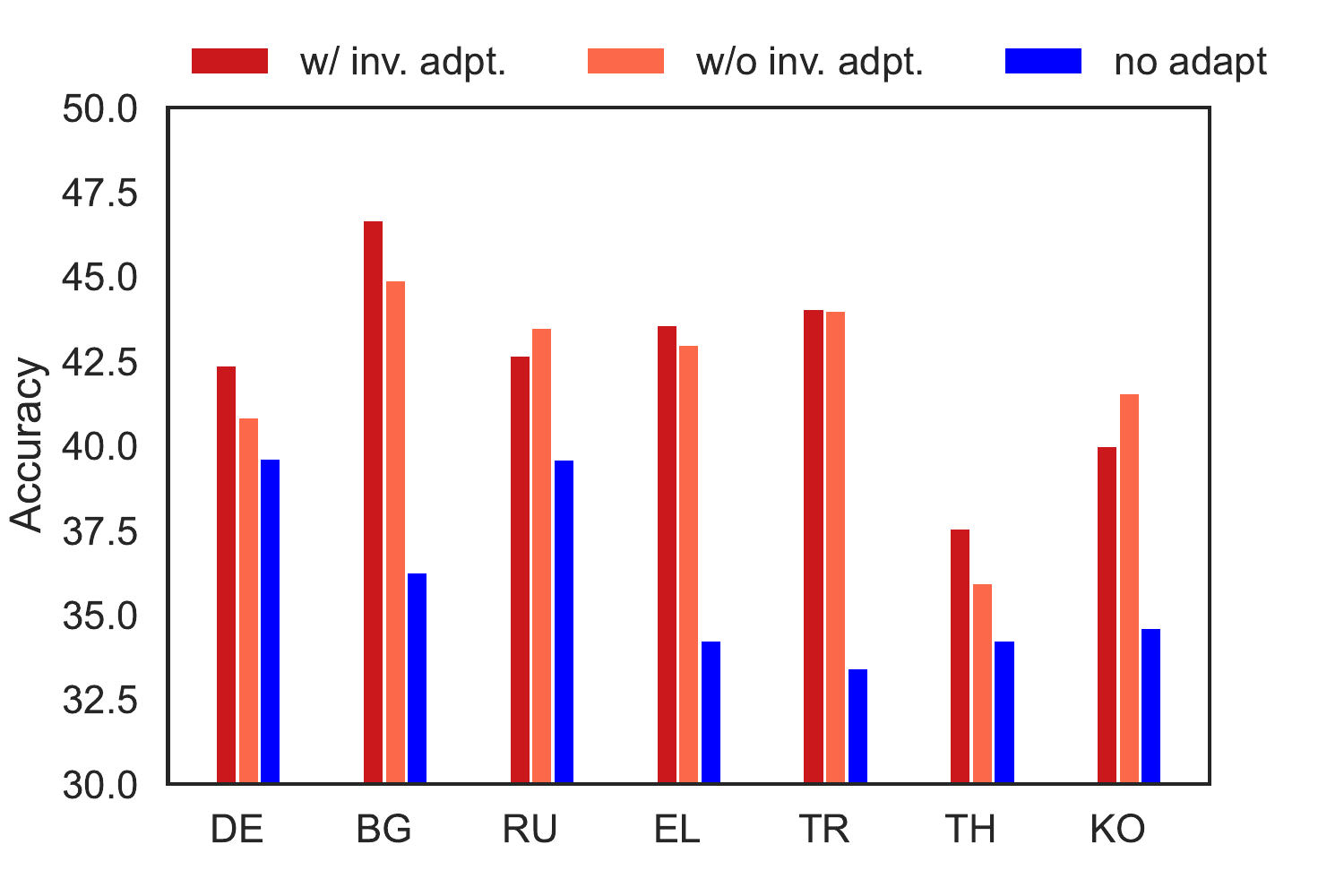}
\centering
\caption{Performance on natural language inference tasks with and without invertible adapters (inv. adpt.) adapting BLOOM's embedding layer. "No adapt" denotes the non-adapted BLOOM model.}
\label{fig:ablation-inv}
\end{figure}

\section{Ablations} \label{app:ablations}
\paragraph{Invertible Adapters} We analyzed the performance of MAD-X with and without invertible adapters, which are used to adapt the embedding layer of BLOOM-3b, on prompting for natural language inference tasks. Figure~\ref{fig:ablation-inv} shows that invertible adapters only improve performance for German, Bulgarian, and Turkish. This implies that the prompting performance gain from language adaptation mainly results from adapting the Transformer blocks.

\begin{figure}[ht]
\includegraphics[width=7.5cm]{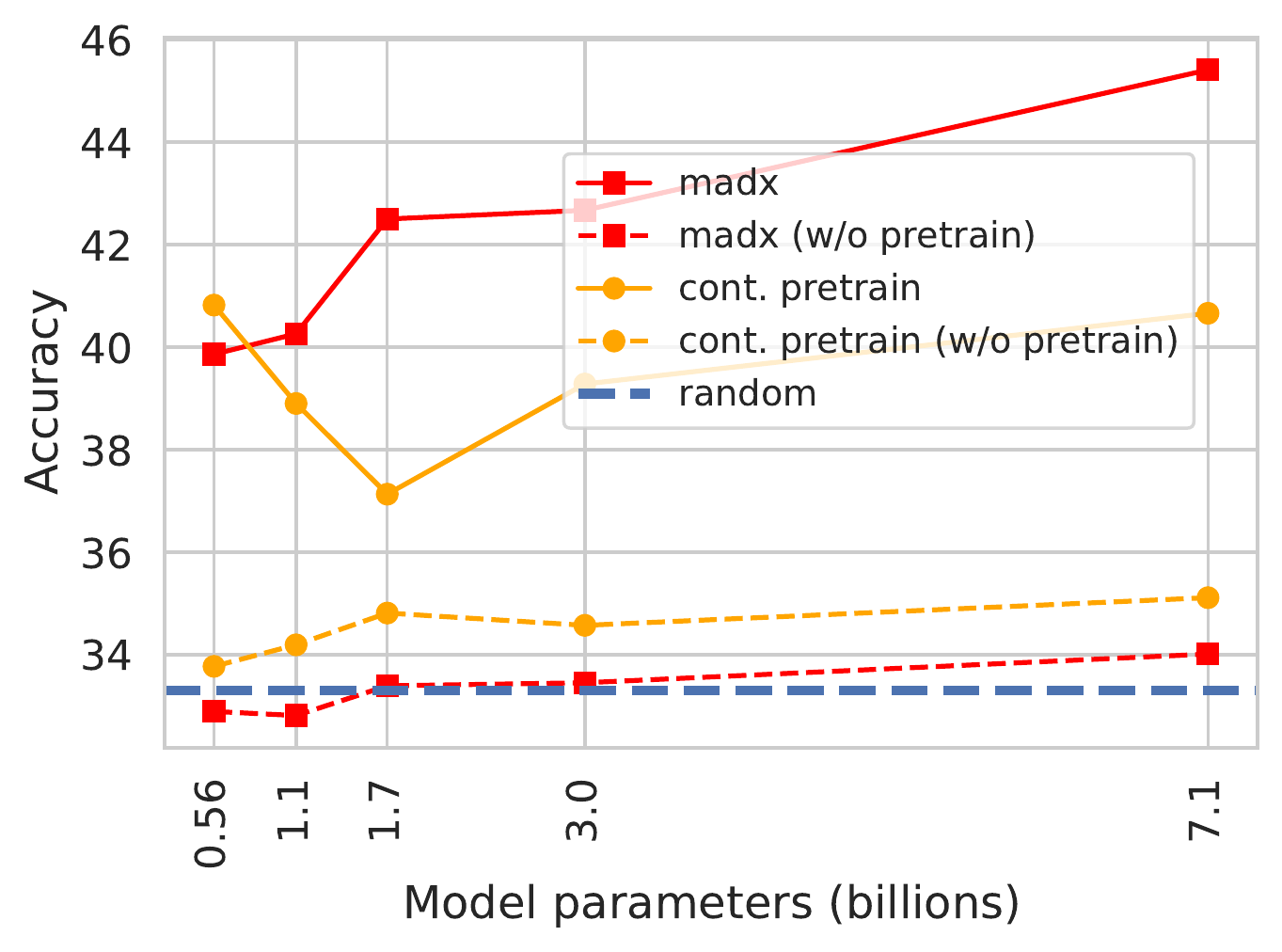}
\centering
\caption{XNLI RU performance with and without pretraining of BLOOM.}
\label{fig:ablation-reinit}
\end{figure}

\paragraph{Model Pretraining} We also performed language adaptation with continued pretraining and MAD-X language adapters on a randomly initialized BLOOM. Figure~\ref{fig:ablation-reinit} shows that, without pretraining, the adapted BLOOM model behaves like a random classifier on the XNLI task. Our results confirm that knowledge transfer takes place during language adaptation of pretrained models. 

\begin{table}[ht]
\small
    \centering
    \begin{tabular}{lcr}
        \toprule
        Languages & Number of Samples & Number of Tokens \\
        \midrule
        BG & 100K & 120M \\
        DE & 100K & 75M \\
        EL & 100K & 160M \\
        GN & 30K & 1M \\
        KO & 100K & 155M \\
        RU & 100K & 140M \\
        RU & 10K & 14M \\
        RU & 1K & 1.4M \\
        TH & 100K & 160M \\
        TR & 100K & 90M \\
        TR & 10K & 9M \\
        TR & 1K & 0.9M \\
        \bottomrule
    \end{tabular}
    \caption{Number of byte-level tokens in the randomly sampled OSCAR data used for language adaptation. Guarani only has 30K samples, fully taken from \citeposs{chiruzzo-etal-2022-jojajovai} corpora. }
    \label{tab:100K-num-tokens}
\end{table}

\section{Catastrophic Forgetting} \label{app:catastrophic-forgetting}
We observe that continued pretraining leads to catastrophic forgetting of seen languages when we evaluated adapted BLOOM on the English XNLI task (Figure~\ref{fig:xnli-en}). 

\begin{figure}[ht]
\includegraphics[width=7.5cm]{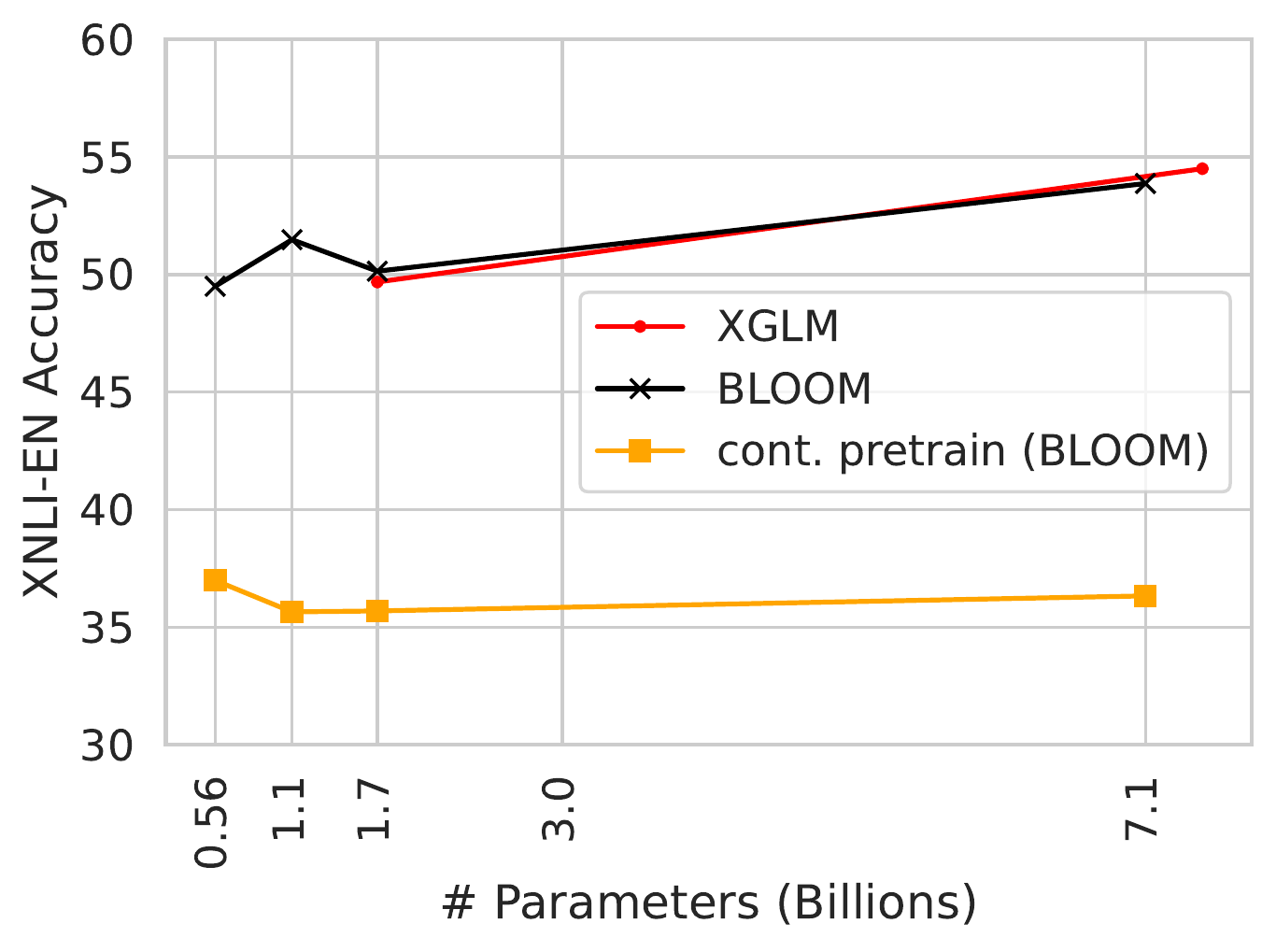}
\centering
\caption{Continued pretraining causes catastrophic forgetting on English, regardless of model sizes.}
\label{fig:xnli-en}
\end{figure}

\section{Pretraining Languages Existing in BLOOM} 
Table~\ref{tab:seen-lang} shows the distribution of natural and programming languages in the ROOTS pretraining data \cite{scao-2022-bloom,laurencon-2022-roots}.

\label{app:seen-lang}
\begin{table}[ht]
\small
    \centering
    \begin{tabular}{lc}
        \toprule
        Language & Proportion (\%) \\
        \midrule
        English & 30.04 \\
        Simplified Chinese & 16.2 \\
        Traditional Chinese & 0.05 \\
        French & 12.9 \\
        Arabic & 4.6 \\
        Basque & 0.15 \\
        Catalan & 1.1\\
        Indonesian & 1.2 \\
        Portuguese & 4.9 \\
        Spanish & 10.8 \\
        Vietnamese & 2.7 \\
        Chitumbuka	& 0.00002 \\
        Assamese & 0.01 \\
        Kikuyu & 0.00004 \\
        Odia & 0.04 \\
        Bambara & 0.00004 \\ 
        Gujarati & 0.04 \\
        Akan & 0.00007 \\ 
        Marathi & 0.05 \\
        Xitsonga & 0.00007 \\ 
        Punjabi & 0.05 \\
        Sesotho & 0.00007 \\ 
        Kannada & 0.06 \\
        Chichewa & 0.0001 \\ 
        Nepali & 0.07 \\
        Setswana & 0.0002 \\ 
        Telugu & 0.09 \\ 
        Northern Sotho & 0.0002 \\ 
        Malayalam & 0.10 \\
        Fon & 0.0002 \\ 
        Urdu & 0.10 \\
        Kirundi & 0.0003 \\ 
        Tamil & 0.20 \\
        Wolof & 0.0004 \\ 
        Bengali & 0.50 \\
        Luganda & 0.0004 \\ 
        Lingala & 0.0002 \\
        Hindi & 0.70 \\
        chiShona & 0.001 \\
        isiZulu & 0.001 \\
        Igbo & 0.001 \\
        isiXhosa & 0.001 \\
        Kinyarwanda & 0.003 \\
        Yoruba & 0.006 \\
        Swahili & 0.02 \\
        Code* & 10.8 \\
        \bottomrule
    \end{tabular}
    \caption{Information about the seen languages by BLOOM model.}
    \label{tab:seen-lang}
\end{table}

\section{Post-Hoc Experiments}
\label{app:post-hoc}
\begin{figure}[ht]
\includegraphics[width=8cm]{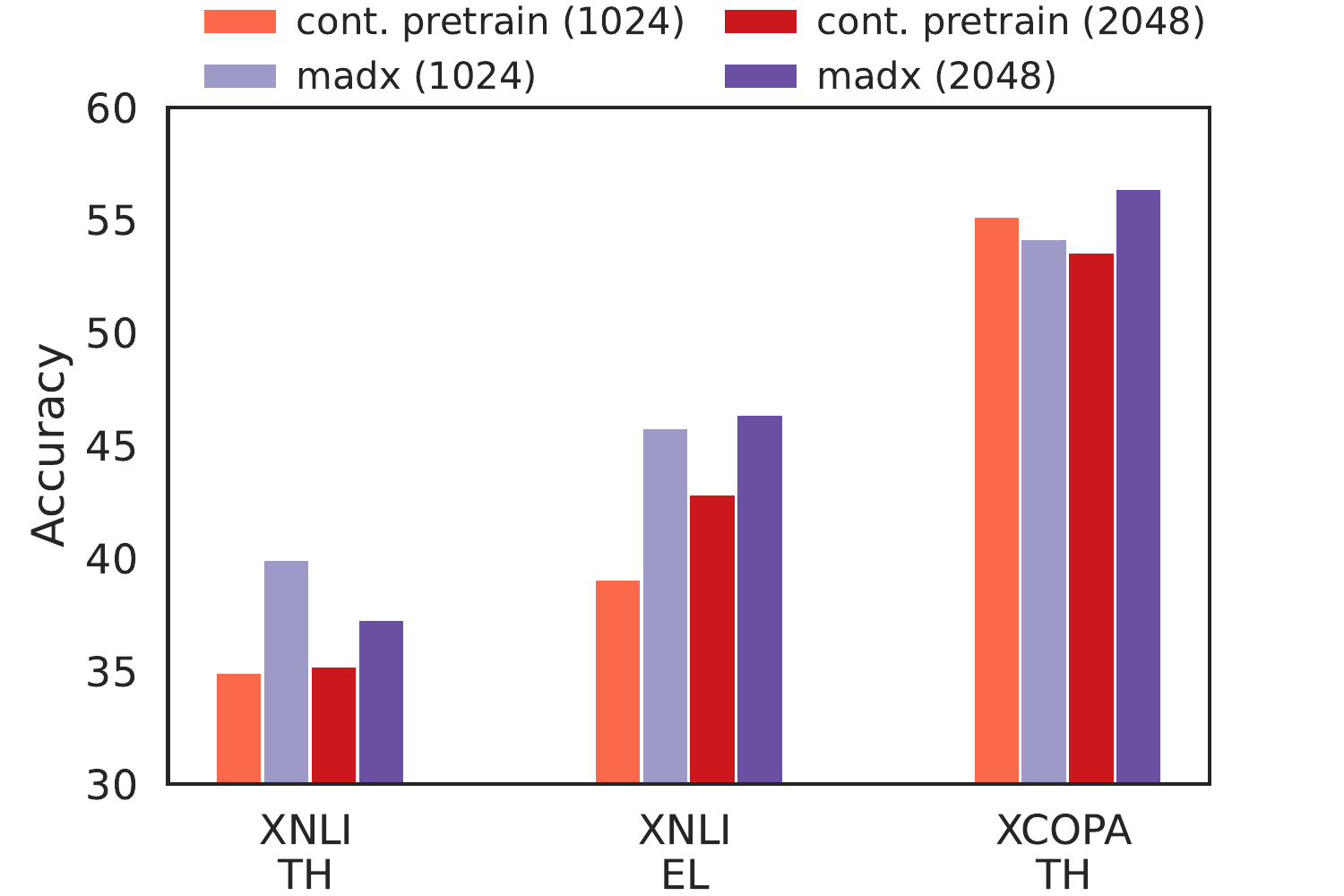}
\centering
\caption{Comparison of prompting performance between sequence lengths of 1024 and 2048 during language adaptation for Thai (TH) and Greek (EL) languages with continued pretraining and MADX adapters. }
\label{fig:post_hoc_seqlen_2048}
\end{figure}

\paragraph{Sequence Lengths of 2048} We adapted BLOOM-7.1B model for Thai and Greek using with the sequence length of 2048 instead of 1024 and training steps of 12500. We picked these two languages because they have the most number of tokens in the 100K samples (see Table~\ref{tab:100K-num-tokens}), and we halved the training steps to maintain the same number of tokens seen during language adaptation since we doubled the sequence length. The rest of the setup follows Section~\ref{sec:lang-adapt-setting}. Figure~\ref{fig:post_hoc_seqlen_2048} shows that adapters-based strategies still outperform continued-pretraining when we use the sequence length of 2048, which is consistent with our results discussed in Section~\ref{sec:result-0-shot-prompting}.

\section{Artifacts}
For the pretrained models used in our study, BLOOM \cite{scao-2022-bloom} and BLOOMZ models \cite{muennighoff2022bloomz} are released under the RAIL license, whereas mGPT \cite{oleh-2022-mgpt} and mT0 \cite{muennighoff2022bloomz} are released under the Apache 2.0 license. XGLM \cite{xi-2021-xglm} is released under the MIT license. 

OSCAR data \citep{ortiz:oscar_2019}, which is used to adapt BLOOM models, are released under the Creative Commons designation CC0 1.0 Universal license. whereas Guarani data \cite{chiruzzo-etal-2022-jojajovai} are released under the MIT license.

XNLI \cite{conneau-etal-2018-xnli} are released under the Attribution-NonCommercial 4.0 International license, KLUE-NLI \cite{park-2021-klue} and AmericasNLI \cite{ebrahimi-etal-2022-americasnli} under the Attribution-ShareAlike 4.0 International license, XCOPA \cite{ponti-etal-2020-xcopa} under the Attribution 4.0 International license, XStoryCloze \cite{xi-2021-xglm} under the MIT license, and PAWS-X \cite{yang-etal-2019-paws} may be freely used for any purpose.

\end{document}